\documentclass{article}
\usepackage{iclr2022_conference,times}

\usepackage{color,framed} % shaded 第1共3
\definecolor{shadecolor}{rgb}{0.92,0.92,0.92}  % shaded 第2共3

\usepackage{xcolor}
\usepackage{hyperref}
\usepackage{amsmath,amssymb,amsfonts,graphicx}
\usepackage{latexsym}
\usepackage{textcomp}
\usepackage{subfigure}
\usepackage{graphics}
\usepackage{epstopdf}
\usepackage{multirow}
\usepackage{bm}
\usepackage{enumerate}
\usepackage{mathdots}
\usepackage{todonotes}
\usepackage{verbatim}
\usepackage{balance}
\usepackage{url}
\usepackage{booktabs}
\usepackage{tabularx}
\usepackage{microtype}
\usepackage{nicefrac}
\usepackage{lipsum}
\usepackage{algorithm}
\usepackage{capt-of}% or \usepackage{caption}
\usepackage{soul}
\usepackage{microtype}
\usepackage{hyperref}
\usepackage{capt-of}
\usepackage{makecell}
\usepackage{wrapfig}
\usepackage{tikzsymbols}
\usepackage{multirow}

\usepackage{capt-of}
\usepackage{makecell}
\usepackage{pifont}   % required for 'ding'
\usepackage{eso-pic,tikz}
\usepackage{listings}

\newcommand{\be}{\begin{equation}}
\newcommand{\ee}{\end{equation}}
\newcommand{\ba}{\begin{align}}
\newcommand{\ea}{\end{align}}

\newcommand{\bi}{\begin{itemize}}
\newcommand{\ei}{\end{itemize}}

\def\ba{{\bf a}}

\def\bx{{\bf x}}
\def\by{{\bf y}}

\def\bA{{\tilde{\bm{A}}}}
\def\bD{{\bf D}}
\def\bE{{\bf E}}
\def\bG{{\bf G}}

\def\bX{{\bf X}}
\def\bY{{\bf Y}}

\def\bW{{\bf W}}

\def\hR{{\mathbb{R}}}
\def\hG{{\mathcal{G}}}
\def\hV{{\mathcal{V}}}

\definecolor{c2}{rgb}{0.9, 0, 1.0} %pink
\definecolor{c3}{rgb}{0, 0, 1}
\definecolor{c4}{rgb}{0, 1, 0}
\definecolor{c5}{rgb}{1, 0.58, 0.0} %orange
\title{
Cold Brew: Distilling Graph Node Representations with Incomplete or Missing Neighborhoods}

\author{Wenqing Zheng, Edward W Huang, Nikhil Rao, Sumeet Katariya, Zhangyang Wang, \\ \textbf{Karthik Subbian}\\
\texttt{\{wenqzhen,ewhuang,nikhilsr,katsumee,wzhangwa,ksubbian\}@amazon.com}
}

\iclrfinalcopy % Uncomment for camera-ready version, but NOT for submission.

\begin{document}
\maketitle

\begin{abstract}
Graph Neural Networks (GNNs) have achieved state-of-the-art performance in node classification, regression, and recommendation tasks. GNNs work well when rich and high-quality connections are available. However, their effectiveness is often jeopardized in many real-world graphs in which node degrees have power-law distributions. The extreme case of this situation, where a node may have no neighbors, is called Strict Cold Start (SCS). SCS forces the prediction to rely completely on the node's own features. We propose \textbf{Cold Brew}, a teacher-student distillation approach to address the SCS and noisy-neighbor challenges for GNNs. We also introduce feature contribution ratio (FCR), a metric to quantify the behavior of inductive GNNs to solve SCS. We experimentally show that FCR disentangles the contributions of different graph data components and helps select the best architecture for SCS generalization. We further demonstrate the superior performance of Cold Brew on several public benchmark and proprietary e-commerce datasets, where many nodes have either very few or noisy connections. Our source code is available at \url{https://github.com/amazon-research/gnn-tail-generalization}.

% \footnote{Our codes are available at: \href{https://github.com/amazon-research/gnn-tail-generalization}{https://github.com/amazon-research/gnn-tail-generalization}}

\end{abstract}

\section{Introduction}
\label{sec:intro}
\vspace{-1em}
\begin{wrapfigure}{r}{0.46\textwidth}
     \vspace{-2em}
    \subfigure{
        \includegraphics[width=2.3in]{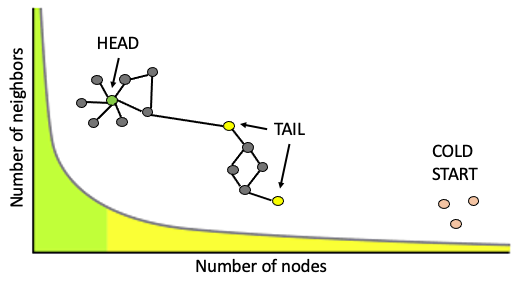}
    }
    \subfigure{
    % \vspace{-3em}
        \includegraphics[width=2.3in]{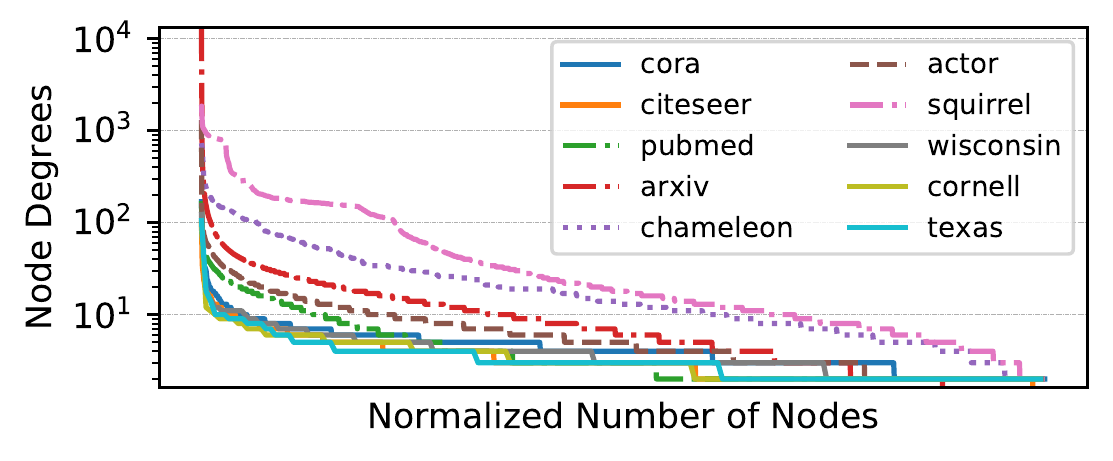}
    }
    \vspace{-0.5em}
    \caption{\small \textbf{Top:} Graph nodes may have a power-law (``long-tail'') connectivity distribution, with a large fraction of nodes (yellow) having few to no neighbors. \textbf{Bottom:} Long-tail distributions in real-world datasets, making modern GNNs fail to generalize to the tail/cold-start nodes.}
    \vspace{-1em}
    \label{fig:motivation}
\end{wrapfigure}

Graph Neural Networks (GNNs) achieve state-of-the-art results across a wide range of tasks such as graph classification, node classification, link prediction, and recommendation~\citep{wu2020comprehensive,goyal2018graph,kherad2020recommendation,shaikh2017recommendation,silva2010graph,zhang2019star}. Most modern GNNs rely on the principle of message passing to aggregate each node's features from its (multi-hop) neighborhood \citep{gcn, GAT, graphsage, gin, sgc, klicpera2018predict}. Therefore, the success of GNNs relies on the presence of dense and high-quality connections. 
%Most if not all modern GNNs \citep{gcn, GAT, graphsage, gin, sgc, klicpera2018predict} benefit from message passing to recursively leverage neighborhood information for node representation learning. Other graph learning methods like the spectral diffusion \citep{joachims2003transductive,chaudhuri2012spectral,zhang2018understanding,wang2021dissecting} and label propagation \citep{CNS,zhu2005semi} also require a non-empty and high-quality neighborhood per node. 
Even inductive GNNs \citet{graphsage} learn a function of the node feature and the node neighborhood, which requires the neighborhood to be present during inference.

%Given the success of modern GNNs, the long tail node degree distribution can cause a unique yet practical challenge. 

A practical barrier for widespread applicability of GNNs  arises from the long-tail node-degree distribution existing in many large-scale real-world graphs. Specifically, the node degree distribution is power law in nature, with a majority of nodes having very few connections \citep{wsdm,ding2021zero, lam2008addressing,lu2020meta}. Figure~\ref{fig:motivation} (top) illustrates a long-tail distribution, accompanied with the statistics of 
several public datasets (bottom). Many information retrieval and recommendation applications face the scenario of \textit{Strict Cold Start} (\textbf{SCS})  \citep{li2019zero,ding2021zero}, wherein some nodes have no edges connected. Predicting for these nodes admittedly is even more challenging than the tail nodes in the graph.
%For instance, in most of the public and proprietary datasets we used in the experimentation we have anywhere between 3\% to 6\% isolated nodes with no neighbors. 
In these cases, existing GNNs fail to perform well due to the sparsity or absence of the neighborhood. 

In this paper, we develop GNN models that have {\it truly inductive} capabilities: one can learn effective node embeddings for ``orphaned'' nodes in a graph. This capability is important to fully realize the potential of large-scale GNN models on modern, industry-scale datasets with very long tails and many orphaned nodes. To this end, we adopt the teacher-student knowledge distillation procedure \citep{yang2021extract, chen2020self} and propose \textbf{Cold Brew} to distill the knowledge of a GNN teacher into a multilayer perceptron (MLP) student. 

The Cold Brew framework addresses two key questions: (1) how we can efficiently distill the teacher's knowledge for the sake of tail and cold-start generalization, and (2) how can a student make use of this knowledge.
% \ks{ this makes me feel like reading goal updates. for 1 we do this. for 2 we do this. Why dont we simplify saying we answer two question (a) and (b) by learning a latent node-wise embedding using knowledge distillation. Then we discuss everything else in detail in contributions. Because after few lines, I see the same content repeated in contributions.} 
We answer these two questions by learning a latent node-wise embedding using knowledge distillation, which both avoids ``over-smoothness'' \citep{oono2020graph,li2018deeper,nt2019revisiting} and discovers latent neighborhoods, which are missing for the SCS nodes. 
% \nr{should we say something about what we mean by absent neighborhoods?}
%We discuss everything else in detail in the following sections.
Note that in contrast to traditional knowledge distillation \citep{hinton2015distilling}, our aim is not to train a simpler student model to perform as well as the more complex teacher. Instead, we aim to train a student model that is better than the teacher in terms of generalizing to tail or SCS samples.
% \eddie{maybe specify "on the same dataset"? since "better than teacher" means generalization to different samples}.
% \nr{I think we should clarify that LABELS are not really necessary. We can train the teacher GNN using self supervision. We focus on semi supervised node calssification in this paper but Cold Brew is a lot more general than that.}. 

In addition, to help select the cold-start friendly model architectures, we develop a metric called \textit{Feature Contribution Ratio} (FCR) that quantifies the contribution of node features with respect to the adjacency structure in the dataset for a specific downstream task. FCR indicates the difficulty level in generalizing to tail and cold-start nodes and guides our principled selection of both teacher and student model architectures in Cold Brew. We summarize our key contributions as follows: 

\vspace{-0.5em}
\begin{itemize}
\vspace{-0.1em}
\item To generalize better to tail and SCS nodes, we design the Cold Brew knowledge distillation framework: we enhance the teacher GNN  by appending the node-wise Structural Embedding (SE) to strengthen the teacher's expressiveness, and design a novel mechanism for the MLP student to rediscover the missing ``latent/virtual neighborhoods,'' on which it can perform message passing.

\vspace{-0.1em}
\item We propose Feature Contribution Ratio (FCR), which 
%a new metric for graph datasets that quantifies the contribution of node features w.r.t. the adjacency structure in the dataset for a specific downstream task. 
quantifies the difficulty in generalizing to tail and cold-start nodes. We leverage FCR in a principled ``screening process''  to select the best model architectures for both the GNN teacher and the MLP student. 
% \nr{reward? it means something very specific in ML, so again please be precise}

\vspace{-0.1em}
\item As the existing GNN studies only evaluate on the entire graph and do not explicitly evaluate on head/tail/SCS, we uncover the hidden differences of head/tail/SCS by creating bespoke train/test splits. Extensive experiments on public and proprietary e-commerce graph datasets validate the effectiveness of Cold Brew in tail and cold-start generalization.

%\item To better capture the teacher's knowledge, we design a \model\ MLP student model that pays attention to virtual neighborhood embeddings learned by the GNN teacher model, which behave like the GNN teacher but can generalize to tail and cold start nodes where the GNN teacher model fails.

%show the superiority of both the SE- enhanced GNN over traditional GNN, and the cold-brew MLP over a naive MLP based student baseline learned to mimic the GNN teacher's node representations
\end{itemize}
%\nr{one part that's unclear is whether \model\ is the whole framework, or just the MLP? we should be clear about it. }
\vspace{-0.5em}
\subsection{Problem Setup}
\vspace{-0.5em}
\label{sec:pf}
GNNs effectively learn node representations using two components in graph data: they process {\it node features} through distributed node-wise transformations and process {\it adjacency structure} through localized neighborhood aggregations. For the first component, GNNs apply shared feature transformations to all nodes regardless of the neighborhoods. For the second component, GNNs use permutation-invariant aggregators to collect neighborhood information.

We take the node classification problem in the sequel for the sake of simplicity. All our proposed methods can be easily adapted to other semi-supervised or unsupervised problem settings, which we show in Section \ref{sec:exps}.  We denote the graph data of interest by $\mathcal{G}$ with node set $\mathcal{V}, ~\ |\mathcal{V}| = N$. Each node possesses a $d_{in}-$dimensional feature and a $d_{out}-$dimensional label (either $d_{out}$ classes or a continuous vector in the case of regression). Let $\bX^{0} \in\hR^{N\times d_{in}}$ and $\bY\in\hR^{N\times d_{out}}$ be the matrices of node features and labels, respectively. Let $\mathcal{N}_i$ be the neighborhood of the $i$-th node, $0\leq i < N$. In large-scale graphs, $|\mathcal{N}_i|$ is often small for a (possibly substantial) portion of nodes. We refer to these nodes as \textit{tail nodes}.  Some nodes may have $|\mathcal{N}_i| = 0$, and we refer to these extreme cold start cases as \textit{isolated nodes}. 
% \nr{Lets use isolated since you use that in your results. Or change the tables in the experiments section to say cold start}

A classical GNN learns representations for the $i^{th}$ node at the $l^{th}$ layer as a function of its representation and its neighborhood's representations at the $(l-1)^{th}$ layer:
\begin{equation}
\label{eq:gnn_standard}
x_i^l := f\left( \{x_i^{l-1}\}, \{ x_j^{l-1}\}_{j\in\mathcal{N}_i}  \right)
\end{equation}
where $f(\cdot)$ is a general function that applies node-wise transformation on node $x_i^{l-1}$ and aggregates information of its neighborhood $\{ x_j^{l-1}\}_{j\in\mathcal{N}_i}$ to obtain the final node representation. Given $i$'s input features $x^0_i$ and its neighborhood $\mathcal{N}_i$, one can use \eqref{eq:gnn_standard} to obtain its representation and predict $y_i$, making these models inductive. 

We are interested in improving the performance of these GNNs on a set of tail and cold-start nodes, where $\mathcal{N}_i$ for node $i$ is either unreliable\footnote{For example, a user with only one movie watched or an item with too few purchases.} or absent. In these cases, applying \eqref{eq:gnn_standard} will yield a suboptimal node representation, since $\{ x_j^{l-1}\}_{j\in\mathcal{N}_i} $ will be unreliable or empty at inference time. 

\vspace{-0.5em}
\section{Related Work}
\vspace{-0.5em}
GNNs learn by aggregating neighborhood information to learn node representations \citep{gcn, GAT, graphsage, gin, sgc, klicpera2018predict}. Inductive variants of GNNs such as GraphSAGE \citep{graphsage} require initial node features as well as the neighborhood information of each node to learn the representation. Most works on improving GNNs have focused on learning better aggregation functions, and methods that can work when the neighborhood is absent or noisy have not been sufficiently exploited, except two recent concurrent works \citep{hu2021graph, zhang2021graph}.

In the context of cold start, \citep{wsdm} and \citep{ding2021zero} employ a transfer learning approach. \citep{yang2021extract} proposes a knowledge distillation approach for GNN, while \citep{chen2020self} proposes a self-distillation approach. In all the above cases, the models need full knowledge of the neighbors of the cold-start nodes in question and do not address the case of noisy or missing neighborhoods. Another possible solution is to directly train an MLP that only takes node features. \citep{hu2021graph} proposes to learn graph embeddings with only node-wise MLP, while using contrastive loss to regularize the graph structure. 
% However, in order to train it with contrastive loss, the training process still relies on neighbor information, and we show through experiments that such approach does not generalize well to tail and cold start nodes.
% cut-short

%There are several previous works that deal with the cold start problems, but not strict cold start. For instance, \citep{wsdm} (WSDM) deals with the cold start in graphs, \citep{ding2021zero} deals with cold start in recommendation system: learn from one dataset then transfer to another dataset with non-overlapping items. However, existing works on cold start on graphs still require the existence of neighbors, hence does not apply to strict cold start cases.

%To solve strict cold start problem, one possible solution is knowledge distillation: by training one MLP that distill the knowledge from the GNN, and generalize to strict cold start cases when the GNN fails. \citep{yang2021extract} (KD2) proposes a knowledge distillation for GNN, but the student still requires explicit neighborhood information. \citep{chen2020self} (KD1) proposed the ``self-distillation'', but still requires neighborhood.

Some previous works have studied the relation between node feature similarity and edge connections and how that influences the selection of appropriate graph models. \citep{pei2020geom} proposed the homophily metric that categorizes graphs into assortative and disassortative classes. \citep{wang2021dissecting} dissected the feature propagation steps of linear GCNs from a perspective of continuous graph diffusion and analyzed why linear GCNs fail to benefit from more propagation steps. \citep{liu2020non} further studied the influence of homophily on model selection and proposed a non-local GNN. 
% Compared with the homophily metric, our proposed FCR quantifies the contribution ratio of node features w.r.t. the adjacency structure. With FCR, one can not only assess the loss of performance for orphaned nodes (through the difference of MLP and GNN) but also obtain the most suitable configuration of the GNN model through a parameter-searching procedure when computing FCR.
% cut-short

% , and witnessed the inliness \nr{what does inliness mean?} of FCR and the homophily metric. \nr{need to say how FCR is different and how is it better}

%In the catagorization of graph datasets, several previous works uncovered the attribute of the correlation between node feature similarity and edge connections, and studied its influence on the selection of appropriate graph models. \citep{pei2020geom} proposed the homophily metric that catagorize the graphs into assortative and disassortative classes. \citep{liu2020non} further studied the influence of homophily on model selection, and proposed the non-local GNN. In our work, we proposed FCR, another metric that catagorizes the respective contribution of node feature and adjacency structure, and witnessed the inliness of FCR and the homophily metric.

\vspace{-0.5em}
\section{Strict Cold Start Generalization}
\label{sec:scs}
\vspace{-0.5em}
We now address the problem of generalization to the tail and cold-start nodes, where the neighborhood information is missing/noisy (Section \ref{sec:intro}). A naive baseline is to train an MLP to map node features to labels. However, such a method would disregard all graph information, and we show via our Feature Contribution Ratio and other experimental results that for most assortative graph datasets, the node-wise MLP approach is suboptimal.

The key idea of our framework is the following: the GNN maps node features into a $d$-dimensional embedding space, and since the number of nodes $N$ is usually much bigger than the embedding dimensionality $d$, we end up with an overcomplete set for this space using the embeddings as the basis. This implies the possibility that any node representation can be cast as a linear combination of $K \ll N$ existing node representations. Our aim will be to train a student model that can accurately discover the combination of the best $K$ existing node embeddings of a target isolated node. We call this procedure \textit{latent/virtual neighborhood discovery}, which is equivalent to using MLPs to ``mimic'' the node representations learned by the teacher GNN.

% \nr{Guys I added some intuition based on what's going on in the section below. Please check and correct this if this does not make sense. }

%In this section, we start tackling the {\bf Strict Cold Start (SCS)} problem. In the SCS case, there usually exists a graph, whose nodes are products/queries/users/merchants/web pages/etc. For those non-cold-start nodes, the labels (embeddings for the sorting purpose) are available, meanwhile, there are also a subset of emerging nodes whose labels are not available, and these nodes can be completely isolated from the existing graph, or their existing connections can be both too small and noisy so as to harm the embedding learning of themselves. We assume that all nodes in the graph are drawn from the same latent distribution.

%As discussed in Section \ref{sec:intro}, due to the missing of neighbors, almost all previous graph researches are helpless to improve SCS performance. However, since the isolated nodes are drawn from the same distribution of the rest of connected nodes, models running on SCS nodes will still be benefited by any possibly available neighbor information, or certain strategy that guides the discovery of latent/virtual neighbors. Therefore, a good SCS generalization model will leverage the latent/virtual neighborhood information while only feed node features in the SCS generalization.

We adopt the knowledge distillation procedure \citep{yang2021extract, chen2020self} to improve the quality of the learned embeddings for tail and cold-start nodes. We use a teacher GNN model to embed the nodes onto a low-dimensional manifold by utilizing the graph structure. Then, the goal of the student is to learn a mapping from the node features to this manifold without knowledge of the graph that the teacher has. We further aim to let the student model generalize to SCS cases where the teacher model fails, beyond just mimicking the teacher as standard knowledge distillation does.

\begin{figure}[t]
\vspace{-1em}
    \centering
    \subfigure[The teacher-student knowledge distillation of the \newline Cold Brew framework under the cold-start setting.]{
        \includegraphics[trim={25cm 12cm 20cm 9cm},clip,width=2.8in]{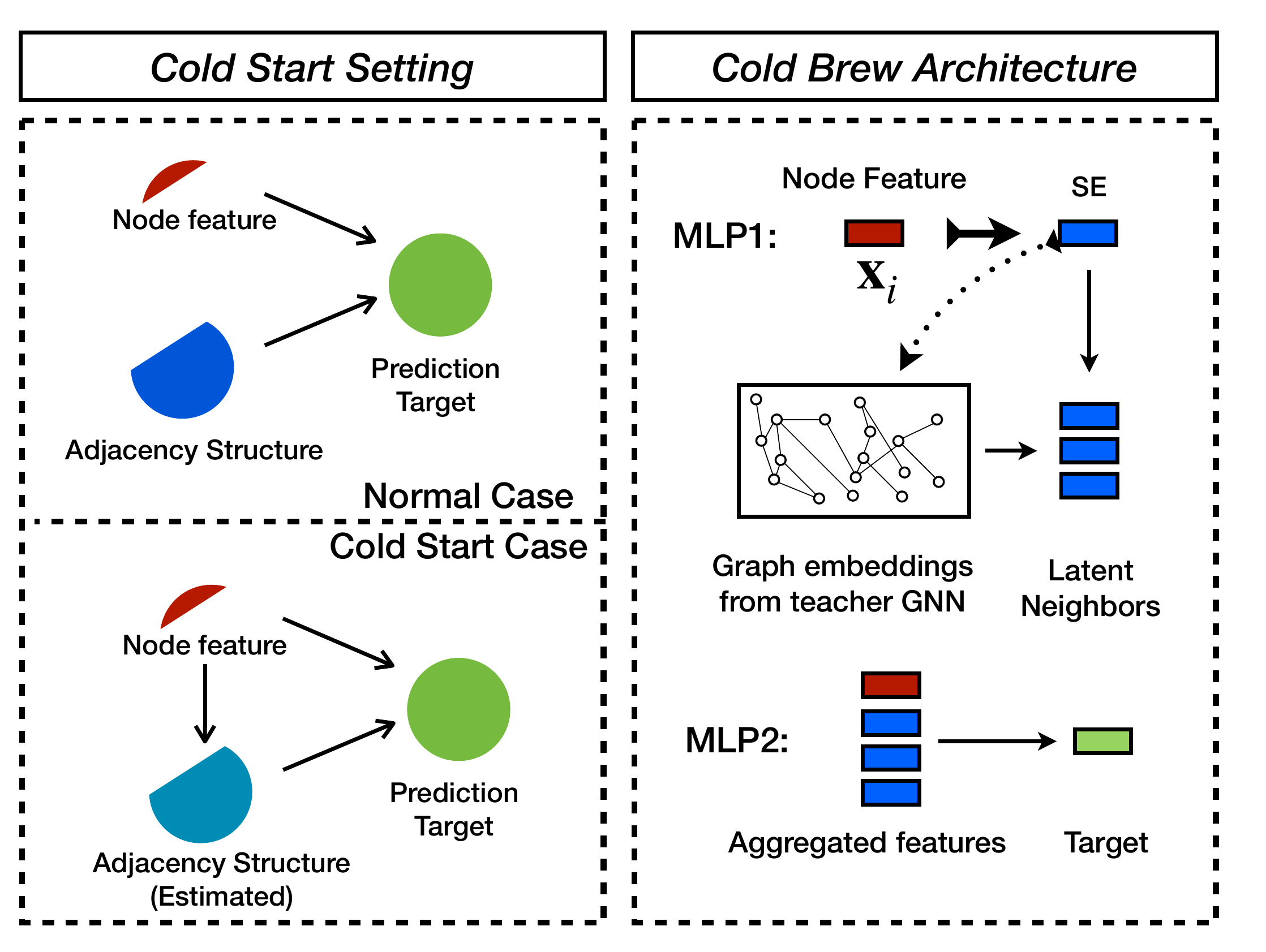}
    }
    \subfigure[Four GNN atomic components in deciding GNN's output, which are used for FCR analysis.]{
        \includegraphics[trim={23cm 7cm 17cm 8cm},clip,width=2.3in]{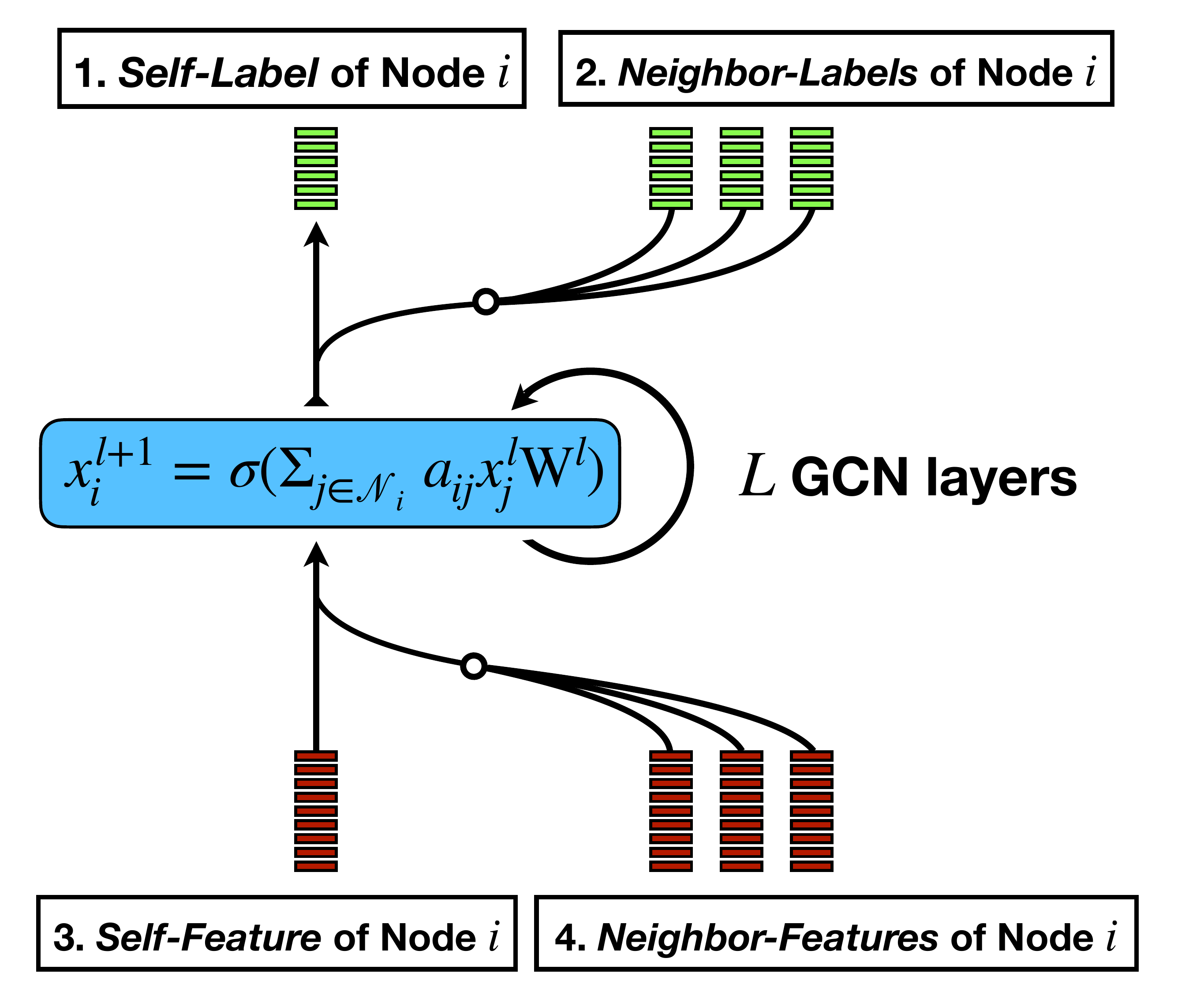}
    }
    \caption{\small \textbf{(a)}: The proposed Cold Brew framework.
    % under the cold start setting: when the adjacency structure is missing (no explicit neighborhood), Cold Brew's student model learns to discover the latent neighborhood, and infer the target embedding from the node feature and the estimated neighbors.
    % cut-short
    In normal case (left upper), GNN relies on both node feature and adjacency structure to make prediction. In cold start case (left lower) when the adjacency structure is missing, the cold brew student model first estimate the adjacency structure, then use both node feature and adjacency structure to make prediction.
    The ``SE'' (right) is the structural embedding learned by Cold Brew's teacher GNN. 
    % \nr{you need to clarify (a) a bit more. what are the arrows on the left side of the figure? why is node feature "smaller" than adjacency structure?}. 
    \textbf{(b)}: Four atomic components deciding the GNN embeddings of node $i$. Our proposed FCR metric disentangles them into two models: the MLP that only considers Part 1 and Part 3, and label propagation that only considers Part 1 and Part 2.}
    % \ks{this plot will make the reader lost. the purpose of the figure is not to read the entire paper and then see the figure. rather the opposite. see the figure, learn the idea and then digest the formulas.}
    \label{fig:illu}
    \vspace{-0.5em}
    \end{figure}
    \vspace{-0.5em}

\subsection{The Teacher Model of Cold Brew: Structural Embedding GNN}
\label{sec:teacher model}
\vspace{-0.5em}
Consider a graph $\mathcal{G}$. For a Graph Convolutional Network with $L$ layers, the $l$-th layer transformation can be written as\footnote{Compared to Equation \eqref{eq:gnn_standard}, multiplication by $\bA$ plays the role of aggregating both $\{x_i\}$ and $\{x_j\}_{i\in\mathcal{N}_i}$.}: $\bX^{(l+1)}=\sigma(\bA\bX^{(l)} \bW^{(l)})$, where $\bA$ is the normalized adjacency matrix, $\bA=\bD^{-1/2}{\bf A}\bD^{-1/2}$, $\bD$ is the diagonal degree matrix, and ${\bf A}$ is the adjacency matrix. $\bX^{(l)}\in\hR^{N\times d_{1}}$ is the node representations in the $l$-th layer, $\bW^{(l)}\in\hR^{d_1\times d_2}$ is the feature transformation matrix, where the values of $d_1/d_2$ depend on layer $l$: $(d_1,d_2)=(d_{in},d_{hidden})$ for $l=0$, $(d_{hidden},d_{hidden})$ for $1\leq l\leq L-2$, and $(d_{hidden},n_{classes})$ for $l=L-1$. $\sigma(\cdot)=$ is the nonlinear functions applied to each layer, (e.g., ReLU). $Norm(\cdot)$ refers to an optional batch or layer normalization.

GNNs typically suffer from oversmoothing \citep{oono2020graph,li2018deeper,nt2019revisiting}, i.e., node representations become too similar to each other. %This is harmful not only for the GNN itself, but also to the student model that looks to mimic the GNN. 
Inspired by the positional encoding in Transformers \citep{vaswani2017attention}, we train the teacher GNN to learn an additional set of node embeddings that can be appended, which we term the {\bf Structural Embedding (SE)}. SE learns to incorporate extra information besides original node features (such as node labels in the case of semi-supervised learning) through gradient backpropagation. The existence of SE avoids the oversmoothing issue in GNNs: the transformations applied to different nodes are no longer the same for every node since the SE of each node is different and participates in the feature transformation. This can be of independent interest to GNN researchers. 

Specifically, for each layer $l,0\leq l\leq L-1$, the Structural Embedding takes the form of a learnable matrix $\bE^{(l)}$, and the SE-GNN layer forward pass can be written as:
\be\label{eq:SE}
\bX^{(l+1)}=\sigma\left(\tilde{\bm{A}}\left(\bX^{(l)} \bW^{(l)}+\bE^{(l)}\right)\right), \bX^{(l)} \in\hR^{N\times d_{1}}, \bW^{(l)} \in\hR^{d_1\times d_{2}}, \bE^{(l)} \in\hR^{N\times d_{2}}
\ee
\paragraph{Remark 1:}Note that SE is not the same as the bias term in traditional feature transformation $\bX^{(l+1)}=\sigma\left(\tilde{\bm{A}}\left(\bX^{(l)} \bW^{(l)}+{\bf b}^{(l)}\right)\right)$; in the bias ${\bf b}\in\hR^{N\times d_{2}}$, the rows are copied/shared across all nodes. In contrast, we have a different structural embedding for every node. 
\paragraph{Remark 2:} SE is also unlike traditional label propagation (LP)~\citep{iscen2019label,wang2020unifying,huang2020combining}. LP encodes label information through iterating $\bE^{(t+1)}=(1-\alpha)\bG+\alpha\bA\bE^{(t)}$, where $\bG$ is a  one-hot encoding of ground truth for training node classes and zeros for test nodes, and $0<\alpha<1$ is the portion of mixture at each iteration. 

SE-GNN enables node $i$ to learn to encode the self and neighbors' label information\footnote{This will be inferred in the case of missing labels.} into its own node embedding through $\bA$. We use the Graph Convolutional Networks \citep{gcn}, combined with other building blocks proposed in recent literature including: (1) initial/dense/jumping connections, and (2) batch/pair/node/group normalization as the backbone of Cold Brew's teacher GNN. More details are described in Appendix \ref{appendix:search space}. We also apply a regularization term to the loss function, yielding the following loss function:
\be
loss=CE(\bX^{(L)}_{train},\bY_{train})+\eta\Vert\bE\Vert_2^2
\ee
where $\bX^{(L)}_{train}$ is the model's embedding at the $L$-th layer, $CE(\bX^{(L)}_{train},\bY_{train})$ is the Cross Entropy loss between the model output $\bX^{(L)}_{train}$ and the ground truth $\bY$ on the training set, and $\eta$ is a regularization coefficient (grid-searched for different datasets in practice). The Cross-Entropy loss can be replaced by any other appropriate loss depending on the task.

\vspace{-0.5em}
\subsection{The Student MLP Model of Cold Brew}
\label{sec:student model}
\vspace{-0.5em}
%Since MLP assumes no input neighborhood, we adopt MLP as the student architecture in \model. 
%In order to improve SCS performance, we aim to ``turn on'' the neighborhood information for the SCS nodes again. 
% To integrate the knowledge the GNN teacher, we find a subset of nodes in the existing connected part of the graph on which the GNN teacher achieves good performance and assign this subset as the latent/virtual neighborhoods of the isolated nodes. 
We design the student to be composed of two MLP modules. Given a target node, the first MLP module imitates the node embeddings generated by the GNN teacher. Next, given any node, we find a set of virtual neighbors of that node from the graph. Finally, a second MLP attends to both the target node and the virtual neighborhood and transforms them into the embeddings of interest. 

%Then, a virtual neighbor nodes discovery process is executed, and a set of virtual neighborhood from the rest connected part of the graph are found, Finally, the second MLP module attends to both the target node and its the discovered neighborhoods, and transform these intermediate results to the node's target embeddings.

Suppose we would like to obtain the embedding of a potentially isolated target node $i$ given only its feature $\bx_i$. 
From the teacher GNN, at each layer $l$, we have access to two sets of node embeddings: $\bX^{(l)}\bW^{(l)}$ and $\bE^{(l)}$. Denote $\bar{\bE}$ as the embeddings that the teacher GNN passes over to the student MLPs. We offer two options for $\bar{\bE}$: it can be the final output of the teacher GNN (in this case, $\bar{\bE}\in\hR^{N\times d_{out}}:= \bX^{(L)}$), or it can be the concatenation of all intermediate results of the teacher GNN, similar to \citep{romero2014fitnets}: $\bar{\bE}\in\hR^{N\times (d_{hidden}*(L-1)+d_{out})}:=\bX^{(L)}\bigcup_{l=0}^{L-1}(\bE^{(l)} + \bX^{(l)}\bW^{(l)})$, where $\bigcup$ is the concatenation of matrices at the feature dimension (second dimension). $\bar{\bE}$ acts as the target for the first MLP and also the input to the second MLP. 

The first MLP learns a mapping from the input node features $\bX^{(0)}$ to $\bar{\bE}$, i.e., for node $i$, $\hat{\bm{e}_i} = \xi_1(\bx^{(0)}_i)$, where $\hat{\bm{e}_i}$ is trained with supervised learning to reproduce $\bar{\bE}[i,:]$. 
Then, we discover the virtual neighborhood by applying an attention-based aggregation of the existing embeddings in the graph before linearly combining them:
\be\label{eq:LEMLP}
\tilde{\bm{e}_i}=softmax(\Theta_K(\hat{\bm{e}_i}\bar{\bE}^\top))\bar{\bE}
\ee
where $\Theta_K(\cdot)$ is the top-$K$ hard thresholding operator: for  $z\in\hR^{1\times N}$: $[\Theta_K(z)]_j=z_j$  if $z_j$ is among the top-$K$ largest elements of $z$, and $\Theta_K(z)_j=-\infty$ otherwise. Finally, the second MLP learns a mapping $\xi_2: \left[\bx_i, \tilde{\bm{e}_i}\right]\to\by_i$, where $\by_i=\bY[i,:]$ is the ground truth for node $i$.

Equation \eqref{eq:LEMLP} first selects $K$ nodes from the $N$ nodes that the teacher GNN was trained on via the hard thresholding operator. $\tilde{\bm{e}_j}$ is then a linear combination of $K$ node\footnote{We abuse terminology here since $\bm{E}$ contains node and structural embeddings from multiple layers.} embeddings. Thus, every sample whether or not seen previously while training the GNN can be represented as a linear combination of these representations. The MLP $\xi_2 (\cdot)$ maps this representation to the final target of interest. Thus, we decompose every node embedding as a linear combination of an (overcomplete) basis. 

% \nr{Guys I tried to add an explanation of what that equation above is doing. Please makea. pass and let me know if this makes sense. }

% During the replacement step, the noisy information in the new node feature was ``denoised'' by changing the estimated profile with the existing profiles in the graph. \nr{is there a proof or a citation for this claim?}
%After the replacement, 

% The second MLP is then applied to map the transformed representations and the original node feature to the learning target: $\by_j = \xi_2([\bx_j, \tilde{\bm{e}_j}])$. 

The training of $\xi_1(\cdot)$ occurs by minimizing the mean squared error over the non-isolated nodes in the graph (mimicking the teacher's embeddings), and the training of $\xi_2(\cdot)$ occurs by minimizing the cross entropy (for the node classification task) or mean squared error (for the node regression task) on the training split of the tail and isolated part of the graph.
An illustration of SE-MLP's inference procedure for the isolated nodes is shown in Figure~\ref{fig:illu}.
When the number of nodes is large, the ranking procedure involved in $\Theta_K(\cdot)$ can be precomputed after training the first part and before training the second part.

\vspace{-0.5em}
\subsection{Model Interpretation From A Label Smoothing Perspective}
\vspace{-0.5em}
We quote Theorem 1 in \citep{wang2020unifying}: \textit{Suppose that the latent ground-truth mapping from node features to node labels is differentiable and L-Lipschitz. If the edge weights $a_{ij}$ approximately smooth $\bx_i$ over its immediate neighbors with error $\epsilon_i$, i.e., $\bx_i=\frac{1}{d_{ii}}\Sigma_{j\in\mathcal{N} }a_{ij}\bx_j+\epsilon_i$,  then the $a_{ij}$ also approximately smooth $y_i$ to bound within error  $|y_i-\frac{1}{d_{ii}}\Sigma_{j\in\mathcal{N}_i }a_{ij} y_j| \leq L||\epsilon||_2+o(\mathrm{max}_{j\in\mathcal{N}_i}(||\bx_j-\bx_i||_2))$, where $o(\cdot)$ denotes a higher order infinitesimal.}

This theorem indicates that the errors of the label predictions are determined by the difference of the features after neighborhood aggregation: if $\epsilon_i$ is large, then the error in the label prediction is also large, and vice versa. However, with structural embeddings, each node $i$ also learns an independent embedding $\bar{\bE}[:,i]$ during the aggregation, which changes $\frac{1}{d_{ii}}\Sigma_{j\in\mathcal{N} }a_{ij}\bx_j+\epsilon_i$ into $\frac{1}{d_{ii}}\Sigma_{j\in\mathcal{N} }a_{ij}\bx_j+\bar{\bE}[:,i]+\epsilon_i$. Deduced from this theorem, the structural embedding $\bar{\bE}$ is important for the teacher model: it allows higher flexibility and expressiveness in learning the residual difference between nodes, and hence the error $\epsilon_i$ can be lowered if $\bar{\bE}$ is properly learned.

From this theorem, one can also see the necessity of introducing neighborhood aggregations like that of the Cold Brew student model. If one directly applies MLP models without neighborhood aggregation, the $\epsilon_i$ turns out to be non-negligible, leading to higher losses in the label predictions. However, Cold Brew introduces the neighborhood aggregation mechanism so that the second part of the student MLP takes over the aggregation of neighborhood generated by the first MLP. Therefore, Cold Brew eliminates the above residual error even in the absence of the actual neighborhood.
% \wenqing{why add punishment to E, add here.}

\vspace{-0.5em}
\section{Model Selection And Graph Component Disentanglement With Feature Contribution Ratio}
\vspace{-1em}
\label{sec:fir}
% To this end, we propose the FCR metric, which gives a measure of the importance of the neighborhood structure for a particular dataset.
% analyzes the contribution of the {\it adjacency structure} of the graph compared to the {\it node features}. 
%  further to guide the model/hyperparameter selection for our task. 
% Also a metric to select best suitable
%As discussed in section~\ref{sec:intro}, the strict cold start problem is a challenging task for graph models, since one of the two components that GNNs extensively rely on -- the {\it adjacency structure} -- is absent, and almost all GNNs will fail in this case. Therefore, before we build new models to generalize to the isolated nodes, we first use the below established measurement to analyze the contribution ratio of the {\it node feature} over the {\it adjacency structure}.
We now discuss Feature Contribution Ratio (FCR), a metric to quantify the difficulty of learning representations under the truly inductive cold-start case, and a hyperparameter optimization approach to select the best suitable model architecture that helps tail and cold-start generalization.

As conceptually illustrated in Figure~\ref{fig:illu}, there are four atomic components contributing to the learned embedding of node $i$ in the graph: 1. the label of $i$ ({\it self-label}); 2. the label of neighbors of $i$ ({\it neighbor-labels}); 3. the features of $i$ ({\it self-feature}); 4. the features of neighbors of  $i$ ({\it neighbor-features}). To quantize the SCS generalization difficulty, we first divide these four components into two submodules to disentangle the contributions of the {\it node features} with respect to the {\it adjacency structure} of the graph dataset. Then, we quantize it based on the assumption that the SCS generalization difficulty is proportional to the contribution ratio of the {\it node features}.

%In order to disentangle the graph dataset into two components (the {\it node feature} and the {\it adjacency structure}), and independently measure the contribution of each component to the overall achievable performance by SOTA GNNs, we %need to appropriately  divide the four atomic components, build two submodules of GNNs out of them, and finally compute the best achievable performance for each submodule. 

We posit that a submodule that learns accurate node representations must include the node's (self) label, so that training can be performed via backpropagation. What remains is to use the label with other atomic components to construct two specialized models that each make use of only the node features or the adjacency structure. For the first submodule, we build an MLP that maps the self-features to self-labels, ignoring any neighborhood information present in the dataset. For the second submodule, we adopt a Label Propagation (LP) method \citep{CNS}\footnote{We ignore the node features and use the label logits as explained in \citep{CNS}.} to learn representations from self- and neighbor-labels. This model ignores any node feature information.

%What remains is to use the {\it self label} with other atomic components to construct two specialized models, and make them account for the {\it node feature} and the {\it adjacency structure} of the graph dataset. The model for {\it node features} can be constructed via MultiLayer Perceptrons (MLPs) that map {\it self-features} to {\it self-labels} (ignoring the neighborhoods). For the second submodule capturing the contribution of the adjacency structure, we combine the {\it self-labels} with {\it neighbor-labels}. We adopt the label propagation reported in one recent SOTA work \citep{CNS} to build the second submodule. \citep{CNS} proposed a two-step {\it outcome correlation} method, where the node-wise {\it outcome} can be genereted either through a node-wise MLP that takes node features, or directly using the label logits that ignore node features. We will adopt the second option for our purpose: our second submodule only takes training node label logits to perform Label Propagation, blind of node features. 

% \nr{why not have it $(1-x)\times 100$ as before? $100 - x$ looks weird no? I'm also confused why there's absolute value in the denominator and not in the numerator?}

% \nr{why is this 1- () ? Also since FCR can be larger than 100, we should not call this a percentage. }

% \wenqing{I think percentage is easy to understand, if the FCR is more than one, e.g. FCR=130\%, it simply means MLP contributes more than 100\% of GNN's performance, Otherwise we say FCR=1.3, does this interpret into the same meaning in human language, what do you think  }

With the above two submodules, we introduce the Feature Contribution Ratio (FCR) that characterizes the relative importance of the node features and the graph structure. Specifically, for graph dataset $\hG$, we define the contribution of a submodule to be the \textit{residual performance} of the submodule compared to a full-fledged GNN (e.g., Equation \eqref{eq:gnn_standard}) using both the node feature as well as the adjacency structure. Denote $z_{MLP}, z_{LP},$ and $z_{GNN}$ as the performance of the MLP submodule, LP submodule, and the full GNN on the test set, respectively. 
If $z_{MLP} \ll z_{GNN}$, then $FCR(\hG)$ is small and the graph structure is important, and noisy or missing neighborhood information will hurt model performance. Based on this intuition, we build SCR as:
\begin{small}
\begin{subequations}\label{eq:fir}
    \begin{align}
\delta_{MLP}=&z_{GNN}-z_{MLP},\quad \delta_{LP} = z_{GNN}-z_{LP} \\
FCR(\hG) = & \begin{cases} \frac{\delta_{LP}}{\delta_{MLP}+\delta_{LP}} \times100\% & z_{MLP} \leq z_{GNN} \\
1 + \frac{|\delta_{MLP}|}{|\delta_{MLP}|+\delta_{LP}}  \times100\% & z_{MLP} > z_{GNN}  \end{cases} 
    \end{align}
    \vspace{-0.5em}
\end{subequations}
\end{small}

\textbf{Interpreting FCR values.} For a particular graph $\hG$, if {$0\% \leq FCR(\hG) < 50\%$}, it means $z_{GNN} > z_{LP} > z_{MLP}$, and the neighborhood information in $\hG$ is mainly responsible for the GNN achieving good performance. If {$50\% \leq FCR(\hG) < 100\%$}, then $z_{GNN} > z_{MLP} > z_{LP}$, and the node features contribute more to the GNN's performance. If {$FCR(\hG) \geq 100\%$}, then $z_{MLP} > z_{GNN} > z_{LP}$, and the node aggregation in GNNs can actually lead to reduced performance compared to pointwise models. This case usually happens for some disassortative graphs, where the majority of neighborhoods hold labels different from that of the center nodes (e.g., as observed by \citep{liu2020non}).

%We use the following assumption: the contribution of a submodule is proportional to the residual performance difference compared to a fully functional GNN model (eg: \eqref{eq:gnn_standard}) using both the node feature as well as the adjacency structure. Given any graph dataset $\hG$,  we compute the \metric by first obtaining the performance of three models: the feature MLP submodule $\bc_{MLP}$, the label propagation submodule $\bc_{LP}$, and a fully functional GNN $\bc_{GNN}$. Then, we record the best achievable accuracy for node classification tasks on the test set as $z_{GNN}$, $z_{MLP}$, and $z_{LP}$. Finally, the \metric for the dataset is given by:

% On one hand, since cold-start for isolated nodes require inferring with only nodes' self feature and may suffer performance drop by degrading from GNN to MLP, FCR offer the exact quantization to this performance degradation by $diff(\bc_{MLP})=z_{GNN}-z_{MLP}$. On the other hand, since the proposed \model is a teacher-student framework and the student performance depend on the teacher, a good teacher model is important to boost the performance of the student. Through the dataset-wise searching of the best performing teacher model, the best suitable teacher can be found, and used for improving the student of \model. \nr{this paragraph is extremely confusing. I think you can just remove this}

% \nr{up until this point in the paper, I still don't see how \metric is useful for \model training. \metric seems to be a measure by which we can determine if \model will actually outperform vanilla GNN or vanilla MLP.  }

\textbf{Integrate FCR as a tool to design teacher and student models.} For some graph datasets and models, the SCS generalization can be challenging without neighborhood information (i.e., $z_{GNN} > z_{LP} > z_{MLP}$). We hence consider FCR as a principled ``screening process'' to select model architectures for both teacher and student that own the best inductive bias for SCS generalization. 

To achieve this, during the computation of FCR, we perform exhaustive grid search of the architectures (residual connection types, normalization, hidden layers, etc.) for the MLP, LP, and GNN modules, and pick the best-performing variant. Detailed definition of the search space can be found in Appendix~\ref{appendix:search space}. We treat this grid search procedure as a special case of architecture selection and hyperparameter optimization for Cold Brew.
%efficient SCS generalization model, we link FCR to our method design as hyperparameter selection in \model: 
We observe that FCR is able to identify the GNN and MLP architectures that are particularly friendly for SCS generalization, improving our method design.

In experiments, we observe that different model configurations are favored by different datasets, and we use the found optimal teacher GNN and student MLP architectures to perform Cold Brew. More detailed discussions are presented in section~\ref{sec:5.3}.

\vspace{-0.5em}
\section{Experiments and Discussion}
\label{sec:exps}
\vspace{-0.5em}
In this section, we first evaluate FCR by training GNNs on several commonly used graph datasets and observing how well they generalize to tail and cold-start nodes. We also compare it to the graph homophiliy metric $\beta$ proposed in \citep{pei2020geom}. Next, we apply Cold Brew to these datasets and compare its generalization ability to baseline graph-based and pointwise MLP models on these datasets. We also show results on proprietary industry datasets. 
\vspace{-0.5em}
\subsection{Datasets and Splits}
\label{sec:datasets}
\vspace{-0.5em}
% \nr{there are 10 in Fig 3, but only 5 in Tables 1 and 2. Are the remaining 5 going to be in appendix?}
% \wenqing{no, in the rest 5 GNN usually underperform MLP since they are disassortative, so no need for cold-brew co-distill; indeed they are not often considered in the experiments of previous papers} \nr{then why not just have 5 in fig 3?}
% \wenqing{The rest are dis-assortative figures, they display different include them}
 We perform experiments on five open-source datasets and four proprietary datasets. The proprietary e-commerce datasets, ``E-comm 1/2/3/4'', refer to graphs subsampled from anonymized logs of an e-commerce store. They are sampled so as to not reflect the actual raw traffic distributions, and results are provided with respect to a baseline model for these datasets. The different number suffixes refer to different product subsets, and the labels indicate product categories that we wish to predict. Node features are text embeddings obtained from a fine-tuned BERT model. We show FCR values for the public datasets. % and exhaustively evaluate the proposed Cold Brew as well as several baselines for five public datasets and the proprietary datasets. 
 The statistics of the datasets are summarized in Table \ref{tab:datasets}.

% %{\bf The special Head/Tail/Isolation splittings}. 
% We create train/validation/test splits of the datasets in order to specifically study the generalization ability of \model to tail and cold-start nodes. %in this work, we split the datasets in a special way. 

%{\bf The special Head/Tail/Isolation splittings}. 
We create training and test splits of the data in order to specifically study the generalization ability of Cold Brew to tail and cold-start nodes.
In the following tables, the {\it head} data corresponds to the top $10\%$ highest-degree nodes in the graph and the subgraph that they induce. We take the data that corresponds to the bottom $10\%$ of the degree distribution, and artifically remove all the edges emanating from these nodes. We then refer to this set of nodes as the {\it isolation} data. The {\it tail} data corresponds to the $10\%$ nodes in the remaining  graph with lowest (non-zero) degree and the subgraph that they induce. All the zero-degree nodes are in the {\it isolation} data. The {\it Overall} data refers to the training/test splits without distinguishing head/tail/isolation. 

% Exact splits of the open source datasets, as well as code to reproduce results are provided in \href{https://github.com/amazon-research/gnn-tail-generalization}{\texttt{https://github.com/amazon-research/gnn-tail-generalization}}

\begin{minipage}[t]{1\columnwidth}
    \begin{center}
    \setlength{\tabcolsep}{1pt}
    \renewcommand{\arraystretch}{1.2}
    \resizebox{0.8\columnwidth}{!}{
    \begin{tabular}{l|c|c|c|c|c|c|c|c|cc}
    % \begin{tabular}[width=1\columnwidth]{l|c|c|c|c|c|c|c|c|cc}
    \toprule[1.5pt]
      &  Cora & Citeseer & Pubmed & Arxiv & Chameleon & E-comm1 & E-comm2 & E-comm3 & E-comm4 \\ 
    \midrule

    Num. of Nodes & 2708 & 3327 & 19717 & 169343 & 2277 & 4918 & 29352 & 319482 & 793194 \\
    Num. of Edges & 13264 & 12431 & 108365 & 2315598 & 65019 & 104753 & 1415646 & 8689910 & 22368070 \\
    Max Degree & 169 & 100 & 172 & 13161 & 733 & 277 & 1721 & 4925 & 12452 \\
    Mean Degree & 4.90 & 3.74 & 5.50 & 13.67 & 28.55 & 21.30 & 48.23 & 27.20 & 28.19\\
    Median Degree & 4 & 3 & 3 & 6 & 13 & 10 & 21 & 15 & 14  \\
    Isolated Nodes $\%$ & 3\% & 3\% & 3\% & 3\% & 3\% & 6\% & 5\% & 5\% & 6\%  \\

    \bottomrule[1.5pt]
    \end{tabular}}
    \captionof{table}{\small The statistics of datasets selected for evaluation.}
    \label{tab:datasets}
    \end{center}
    \end{minipage}

\vspace{-0.5em}
\subsection{FCR Evaluation}
\vspace{-0.5em}
\label{sec:beta}
In Table~\ref{fig:heng}, the top part presents the FCR results together with the homophily metric $\beta$ from \citep{pei2020geom} (Equation \ref{eq:beta}). The bottom part shows the prediction accuracies for the head and the tail nodes. As can be seen from the table, FCR differs among datasets and is negatively correlated with the homophily metric (with Pearson correlation coefficient -0.76). The high absolute correlation value and its negative sign indicate that the more similar the nodes are to their neighborhoods, the more difficult it is to generalize with MLP based models. FCR is thus an indicator of the tail generalization difficulty. Evaluations on more datasets (including the datasets where FCR $>100\%$) are presented in Appendix ~\ref{appendix:all exps}. \begin{small}
\be\label{eq:beta}
\vspace{-1em}
\beta(\hG)=\frac{1}{|\hV|}\sum_{v\in\hV}\frac{\textrm{the number of $v$'s direct neighbors that have the same labels as $v$}}{\textrm{the number of $v$'s directly connected neighbors}}\times100\%
\vspace{-1em}
\ee
\end{small}

\begin{table}[h]
    \centering
     \vspace{-0.5em}
    \resizebox{0.6\columnwidth}{!}{
    \begin{tabular}{c|c  c   c c c cccccc  }
    % \centering
    \toprule[1.5pt]
    % \multirow{2}{*}{\textbf{\makecell[c]{Dataset}}} & \multicolumn{4}{c}{\textbf{\makecell{d}}} \\ %& \multicolumn{2}{c|}{\textbf{\makecell{Grid-Uni}}} & \multicolumn{2}{c}{\textbf{\makecell{NewYork}}}\\
    % \cmidrule(r){2-5}
     & Cora & Citeseer & Pubmed & Arxiv & Chameleon  \\
    % Dataset & GCN best configure & MLP Best configure & LP best configure  \\
       \hline
    GNN  & 86.96& 72.44& 75.96& 71.54& 68.51& \\

    MLP  & 69.02& 56.59& 73.51& 54.89& 58.65& \\
    Label Propagation  &  78.18& 45.00  & 67.8 & 68.26& 41.01\\

    FCR \% & 32.86 \% &  63.39 \% &  76.91\% &  16.45\% &  73.61\% \\

    $\beta(\hG)$ \% & 83\% & 71\% & 79\% & 68\% & 25\% & \\
       \hline
    $head-tail (GNN)$ &  4.44&  23.98&  11.71&   5.9 &   0.24\\
    $head-isolation (GNN)$ & 31.01&  33.09&  15.21&  28.81&   1.55& \\

    \bottomrule[1.5pt]
    \end{tabular}
    }
    \caption{\small Top part: FCR and its components. The $\beta$ metric is added as a reference. Bottom part: the performance difference of GNN on the head/tail and head/isolation splits. Here, the ``tail/isolation'' means the 10\% least connected, and isolated nodes in the graph.}
    % \ks{explain the head-tail and head-isolation in text and caption}
    \label{fig:heng}
    \vspace{-1em}
    \end{table}

% \begin{minipage}[b]{0.37\columnwidth}
% \begin{center}
% \setlength{\tabcolsep}{1pt}
% \renewcommand{\arraystretch}{1.2}
% \resizebox{5cm}{!}{
% \begin{tabular}{lccccc}
% \toprule
% Attribute & Value \\
% \midrule
% Num. Nodes & 10MM \\
% Num. Edges & 100MM \\
% Num. Node-Types & 2 \\
% Node Types & $\mathrm{Query\ keywords, product\ title}$ \\
% Node Feature Type & $\mathrm{Texts}$ \\
% Edge Types & $\mathrm{Exact\ \&\ Inexact\ product}$ \\
% Isolated Nodes Ratio & 30.0\% \\
% Mean Node Degree & 10 \\
% Median Node Degree & 25 \\
% \bottomrule
% \end{tabular}}

% \captionof{table}{The statistics of the E-commerce dataset.}
% \label{tab:E-commerce}
% \end{center}
% \end{minipage}

\vspace{-1em}
\subsection{Experimental Results on Tail Generalization With Cold Brew}
\label{sec:5.3}
\vspace{-1em}
In Table~\ref{tab:proposed}, we present the performance of  Cold Brew together with baselines on the tail and the isolation splits, across several different datasets. All the models in the table are evaluated on the training data, and are evaluated on the tail or isolation splits discussed in section~\ref{sec:datasets}. 
% For the isolation split, the GNNs are evaluated with only self-loop (all other edges are non-existent/removed). 
In Table~\ref{tab:proposed} {\it GCN} refers to the the best configuration found through FCR-guided grid search (check Appendix~\ref{appendix:search space} for details), without Structural Embedding. Correspondingly, {\it GCN + SE} refers to the best FCR-guided configuration with Structural Embedding, which is the default teacher model of Cold Brew. {\it GraphSAGE} refers to \citep{graphsage}, {\it Simple MLP} refers to a simple node-wise MLP that has two hidden layers with 128 hidden dimensions, and {\it GraphMLP} refers to \citep{hu2021graph}. The results for the e-commerce datasets are presented as relative improvements to the baseline (each value is the difference with respect to the value of the {\it GCN 2 layers} on same dataset of the same split). We do not disclose the absolute numbers due to proprietary reasons.

As shown in Table~\ref{tab:proposed}, Cold Brew's student MLP improves accuracy on isolated nodes by up to +11\% on the e-commerce datasets and +2.4\% on the open-source datasets. Cold Brew's student model handles isolated nodes better, and the teacher GNN also achieves better performance on the tail split compared to all other models. 
Especially when compared with GraphMLP, Cold Brew's student MLP consistently performs better. This can be explained from their different mechanisms: GraphMLP encodes graph knowledge implicitly in the learned weights, while Cold Brew explicitly attends to neighborhoods even when they are absent. More detailed comparisons can be found in Appendix \ref{appendix:all exps}.

% Due to the structural embedding, Cold Brew's teacher GNN efficiently captures the neighborhood structure information and passes them to the student MLP. This helps the student to distill graph knowledge from the teacher, rather than being blind to the graph if used in isolation. 

\begin{table}[h]
     \vspace{-0.5em}
    \centering
    \resizebox{0.85\textwidth}{!}{
    \begin{tabular}{l|lc|ccccc|cccccccccc}
    \toprule[1.5pt]
    \multirow{2}{*}{Splits} & \multicolumn{2}{c}{\multirow{2}{*}{Metrics/Models}} & \multicolumn{5}{|c|}{\bf Open-Source Datasets} & \multicolumn{4}{c}{\bf Proprietary Datasets} &  \\
    \cmidrule(r){4-8} \cmidrule(r){9-12}
    & & & Cora & Citeseer & Pubmed & Arxiv & Chameleon & E-comm1 & E-comm2  & E-comm3 & E-comm4 \\ 
    \midrule

    % __________________________________ exps here! __________________________________

      \multirow{7}{*}{Isolation}
     & \multirow{2}{*}{GNNs}  &  GCN 2 layers   & 58.02 & 47.09 & 71.50 & 44.51 & 57.28 & $-$ & $-$ & $-$ & $-$\\
     &   & GraphSAGE & 66.02 & 51.46 & 69.87 & 47.32 & 59.83 & +3.89 & +4.81 & +5.24 & +0.52\\
     \cmidrule(r){2-2}
     &  \multirow{2}{*}{MLPs}  &  Simple MLP  & 68.40 & \textbf{53.26} & 65.84 & 51.03 & 60.76 & +5.89 & +9.85 & +5.83 & +6.42\\
     &   &  GraphMLP  & 65.00 & 52.82 & 71.22 & 51.10 & \textbf{63.54 } & +6.27 & +9.46 & \textbf{+5.99 } & +7.37\\
     \cmidrule(r){2-2}
     &   \multirow{2}{*}{Cold Brew} &   GCN + SE 2 layers   & 58.37 & 47.78 & \textbf{73.85} & 45.20 & 60.13 & +0.27 & +0.76 & -0.50 & +1.22\\
     &   &  Student MLP  & \textbf{69.62} & 53.17 & 72.33 & \textbf{52.36} & 62.28 & \textbf{+7.56} & \textbf{+11.09} & +5.64 & \textbf{+9.05}\\

    \midrule  
     \multirow{7}{*}{Tail}

     & \multirow{2}{*}{GNNs}  &  GCN 2 layers   & 84.54 & \textbf{56.51} & 74.95 & 67.74 & 58.33 & $-$ & $-$ & $-$ & $-$\\
     &   & GraphSAGE & 82.82 & 52.77 & 73.07 & 63.23 & 61.26 & -3.82 & -3.07 & -2.87 & -6.42\\
     \cmidrule(r){2-2}
     &  \multirow{2}{*}{MLPs} &  Simple MLP  & 70.76 & 54.85 & 67.21 & 52.14 & 50.12 & -0.37 & +1.74 & -0.13 & -0.45\\
     &   &  GraphMLP  & 70.09 & 55.56 & 71.45 & 52.40 & 52.84 & -0.33 & +1.64 & \textbf{+1.27 }& +0.80\\
     \cmidrule(r){2-2}
     &  \multirow{2}{*}{Cold Brew}   &   GCN + SE 2 layers   &\textbf{ 84.66 }& 56.32 & \textbf{75.33} & \textbf{68.11} & \textbf{60.80} & \textbf{+0.85} & +0.44 & -0.60 & +1.10\\
     &   &  Student MLP  & 71.80 & 54.88 & 72.54 & 53.24 & 51.36 & +0.32 & \textbf{+3.09 }& -0.18 & \textbf{+2.09}\\

    \bottomrule[1.5pt]
    \end{tabular}}
    \caption{\small The performance comparisons on the isolation and tail splits of different datasets. The full comparisons on head/tail/isolation/overall data are in the Appendix \ref{appendix:all exps}. GCN+SE 2 layers is Cold Brew's teacher model. Cold Brew outperforms GNN and other MLP baselines, and achieves the best performance on the isolation splits as well as some tail splits.}
    % \ks{key insights must go here, lets save time for the reader.}
    \label{tab:proposed}
    \vspace{-1em}
    \end{table}

\begin{center}
\begin{minipage}[t]{0.43\columnwidth}
    \resizebox{0.96\textwidth}{!}{
    \begin{tabular}{l|l|cccccccccccccc}
    \toprule[1.5pt]
    \multirow{2.5}{*}{Splits} & \multicolumn{1}{c}{\multirow{2.5}{*}{Models}} & \multicolumn{4}{|c}{\bf  Datasets}\\ % & \multicolumn{4}{c}{\bf Proprietary Datasets} &  \\
    % Splits &  Metrics/Models & Datasets \\
    \cmidrule(r){3-6}
    & & Cora & Citeseer & Pubmed & E-comm1 \\ 
    \midrule
     \multirow{5}{*}{Isolation}
     & GCN 2 layers & 34.10  & 50.41 &  51.52 & $-$  \\
     \cmidrule(r){2-2}
     & TailGCN   &  36.13  & 51.48  & 51.19 & +2.18 \\
     \cmidrule(r){2-2}
     &  Meta-Tail2Vec & 36.92 & 50.90 & 51.62 & +2.34  \\
     \cmidrule(r){2-2}
     & Cold Brew's MLP  & \textbf{44.59} &  \textbf{55.14} & \textbf{54.82}  & \textbf{+5.39} \\
    \bottomrule[1.5pt]
    \end{tabular}}
    \captionof{table}{\small Link prediction Mean Reciprocal Ranks (MRR) on the isolation data. Note that Cold Brew outperforms baselines specifically built for generalizing to the tail. }
    \label{tab:nov linkp}
\end{minipage}
\begin{minipage}[t]{0.4\columnwidth}
\end{minipage}
\begin{minipage}[t]{0.43\columnwidth} % change width, will not change figure size but change the anchor point of the right starting point (can overlap)
    \resizebox{0.96\textwidth}{!}{
    \begin{tabular}{l|l|cccccccccccccc}
    \toprule[1.5pt]
    \multirow{2.5}{*}{Splits} & \multicolumn{1}{c}{\multirow{2.5}{*}{Models}} & \multicolumn{4}{|c}{\bf  Datasets}\\ % & \multicolumn{4}{c}{\bf Proprietary Datasets} &  \\
    % Splits &  Metrics/Models & Datasets \\
    \cmidrule(r){3-6}
    & & Cora & Citeseer & Pubmed & E-comm1 \\ 
    \midrule

     \multirow{5}{*}{Isolation}
     & GCN 2 layers & 58.02 & 47.09 & 71.50 & $-$ \\
     \cmidrule(r){2-2}
     & TailGCN   &  62.04  & 51.87 & 72.10 &  +3.14 \\
     \cmidrule(r){2-2}
     &  Meta-Tail2Vec & 61.16 & 50.46 & 71.80 & +2.80  \\
     \cmidrule(r){2-2}
     & Cold Brew's MLP  &  \textbf{69.62} & \textbf{53.17} & \textbf{72.33} & \textbf{+7.56} \\

    % \midrule  
    %  \multirow{5}{*}{Tail}
    %  & GCN 2 layers & 84.54 &  & \\
    %  \cmidrule(r){2-2}
    %  & TailGNN   & 58.02 & 47.09 & 71.50 & 44.51 \\
    %  \cmidrule(r){2-2}
    %  &  Meta-Tail2Vec & 66.02 & 51.46 & 69.87 & 47.32  \\
    %  \cmidrule(r){2-2}
    %  & Cold Brew's MLP  & 71.80 & 54.88 & 53.24 & +0.32 \\

    \bottomrule[1.5pt]
    \end{tabular}}
    \captionof{table}{\small Node classification accuracies with other baselines specifically created to generalize to the tail. Cold Brew outperforms these methods when edge data is absent in the graph. }
    \label{tab:nov classi}
\end{minipage}
\end{center}

We also evaluated the link prediction performance by replacing the node classification loss with the link prediction loss. On the manually created isolation split, the model is asked to recover the ground truth edges which are manually removed.
The results are shown in Table \ref{tab:nov linkp}. The baseline models shown in table are TailGCN \citep{vetter1991eta} and Meta-Tail2Vec \citep{liu2020towards}. A comparison over these models on the node classification on the isolation split is provided in Table \ref{tab:nov classi}. As observed from the table \ref{tab:nov linkp} and \ref{tab:nov classi}, Cold Brew outperformed TailGCN and Meta-Tail2Vec on the isolation split, since both TailGCN and Meta-Tail2Vec explicitly are not zero-shot methods and require explicit neighborhood nodes, hence their performance degrades when the neighborhood is empty and padded by zero vectors.

The full performance on other splits are listed in Table \ref{tab:fullexp} in the appendix as a reference. The results across all splits in Table \ref{tab:fullexp} provide evidence for a few phenomena, for example, the high FCR means that the graph structure does not add too much information for the task at hand, and that GNN type models tend to perform better on the head while MLP type models tend to perform better on the tail/isolation splits.
On the other hand, the proposed Structural Embeddings imply a potential to alleviate the over-smoothness \citep{oono2020graph,li2018deeper,nt2019revisiting} and bottleneck \citep{alon2020bottleneck} issues observed in deep GCN models. 
As shown in table Table~\ref{tab:64}, Cold Brew's GCN (GCN + SE) significantly outperformed the traditional GCN on 64 layers: the former has 34\% test accuracy higher on Cora, 23\% higher on Citeseer, and similar on others.

Finally, the improvement over isolation and tail splits (especially the isolation split) comes with a cost: we observed a performance drop for the student MLP model on the head and several other datasets' tail splits, compared with the naive GCN model. 
However, Cold Brew specifically targets the challenging strict cold start issues, as a new compelling alternative for in these cases. Meanwhile in the non-cold-start cases, the traditional GCN models can still be used to obtain good performance. Note that even on the head splits, the proposed GNN teacher model of Cold Brew still outperformed traditional GNN models. We hence consider as promising future work to adaptively switch between using Cold Brew teacher and student models, based on the current node connectivity degree.
% \nr{is tailedness a word?}

\begin{table}[h]
    \centering
    \resizebox{0.8\textwidth}{!}{
    \begin{tabular}{l|l|ccccc|cccccccccc}
    \toprule[1.5pt]
    \multirow{2}{*}{Splits} & \multirow{2}{*}{Metrics/Models} & \multicolumn{5}{c|}{\bf Open-Source Datasets} & \multicolumn{4}{c}{\bf Proprietary Datasets} &  \\
    \cmidrule(r){3-7} \cmidrule(r){8-11}
    & & Cora & Citeseer & Pubmed & Arxiv & Chameleon & E-comm1 & E-comm2  & E-comm3 & E-comm4 \\ 
    \midrule

    % \multirow{1}{*}{Metric}

    %  % &  $\beta$ & 98.22 & 83.16 & 76.32 & 66.32 & 80.69 & 75.91 & 89.21 & 96.87 & 69.72 & 97.44 & 52.32 \\
    %  % &  $\beta$ & 98.22 & 83.16 & 76.32 & 66.32 & 80.69 & 75.91 & 89.21 & 96.87 & 69.72  \\

    %  &  FCR & 32\%  & 63\% & 76\%  &  16\%   &  73\%  &  61\%  &  51\%  & 54\%  &  44\%     \\

    % \midrule

    %  exps here! __________________________________

     \multirow{2}{*}{Overall} 
     
      &  GCN 64 layers  & 40.04 & 23.66 & 75.65 & 65.53 & 58.14 & -5.49 & -6.59 & -6.13 & -3.57\\
     & GCN + SE 64 layers & \textbf{74.23} & \textbf{46.80} & \textbf{78.12} & \textbf{69.28} & \textbf{59.88} & \textbf{-1.71 }& \textbf{-2.92} & \textbf{-3.29} & \textbf{-0.06}\\

      \midrule 
     \multirow{2}{*}{Head} 
     
     &  GCN 64 layers  & 46.46 & 49.84 & 85.89 & 67.53 & 67.16 & -5.60 & -6.24 & -6.05 & -3.16\\
     & GCN + SE 64 layers & \textbf{87.38} &\textbf{ 71.18 }& \textbf{86.81} & \textbf{71.35} & \textbf{69.63} & \textbf{-1.78} &\textbf{ -2.17} &\textbf{ -2.79} & \textbf{-0.35} \\

    \midrule  
     \multirow{2}{*}{Tail}

     &  GCN 64 layers  & 45.14 & 24.42 & 71.89 & 63.91 & 56.48 & -3.85 & -3.62 & -3.84 &\textbf{ -1.14}\\
     & GCN + SE 64 layers & \textbf{79.56} & \textbf{36.52} & \textbf{74.88} & \textbf{65.19} & \textbf{61.73} & \textbf{-2.42} & \textbf{-2.52} &\textbf{ -3.68} & -1.23\\

     \midrule 
     \multirow{2}{*}{Isolation}
     &  GCN 64 layers  & 39.97 & 22.12 & 68.57 & 40.03 & 57.60 & -4.66 & -4.63 & -4.93 & \textbf{-1.89} \\
     & GCN + SE 64 layers & \textbf{40.33} & \textbf{24.53} & \textbf{71.22} &\textbf{ 41.18} & \textbf{60.13} & \textbf{-3.08} & \textbf{-3.02} & \textbf{-4.00} & -2.32\\

    % __________________________________ exps here! __________________________________

    \bottomrule[1.5pt]
    \end{tabular}}
    \caption{\small The comparisons of Cold Brew's GCN and the traditional GCN for deep layers. When the number of layers is large, Cold Brew's GCN retains good performance while the traditional GCN without SE suffers from the ``over-smoothess'' and degrades. Even with shallow layers, Cold Brew's GCN is better than traditional GCN.}
    % \ks{lets save time for the reader, share the key insights here from the table}
    \label{tab:64}
    \vspace{-1em}
    \end{table}

% The comparisons of Cold Brew's GCN and the traditional GCN for deep layers are put in appendix~\ref{appendix:deep}

% (also recall the disassortative Cornell and Wisconsin datasets shown in Fig.~\ref{fig:heng}). 

% \nr{You need to call out that \model is better in Tail and Isolation split. \model GCN is usually the best in the tail and \model MLP lets us learn representations that achieve SOTA in Isolation. }

\vspace{-1em}
\section{Conclusion}
\vspace{-1em}
In this paper, we studied the problem of generalizing GNNs to the tail and strict cold start nodes, whose neighborhood information is either sparse/noisy or completely missing. We proposed a teacher-student knowledge distillation procedure to better generalize to the isolated nodes. We added an independent set of structural embeddings in GNN layers to alleviate node over-smoothness, and also proposed a virtual neighbor discovery step for the student model to attend to latent neighborhoods.  We additionally present the FCR metric to quantify the difficulty of truly inductive representation learning and to optimize our model architecture design. Experiments demonstrated the consistently superior performance of our proposed framework on both public benchmarks and proprietary datasets.

\bibliography{refs}
\bibliographystyle{iclr2022_conference}
% \bibliographystyle{unsrt}

% \newpage
\appendix

\section{More illustrations}
\label{sec:illu}

\begin{figure}[!h]
\begin{center}
% crop caijian jianqie trim={<left> <lower> <right> <upper>}
\centerline{\includegraphics[trim={1cm 520 50 30},clip,width=\columnwidth]{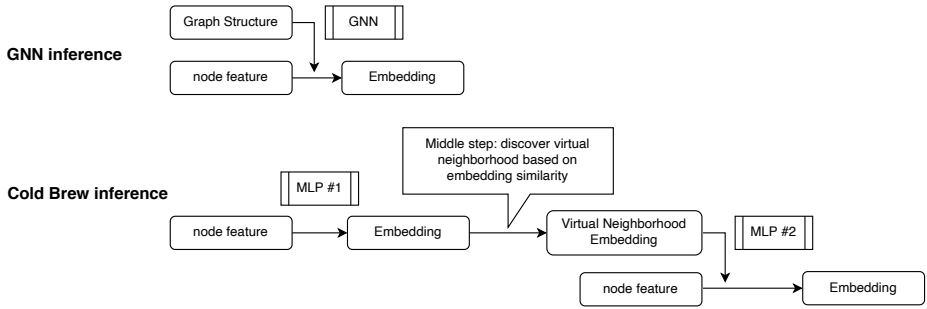}}
\caption{Inference procedure illustration for GNN and Cold Brew.}
\label{fig:flowillu}
\end{center}
\end{figure}

The more detailed inference procedures for GNN and Cold Brew are illustrated in Figure~\ref{fig:flowillu}.

\section{Search Space Details}
\label{appendix:search space}

In computing FCR, we include a search space of model hyperparameters for GNN, MLP, and LP in order to find the best suitable configurations for distillation.

For the GNN model, we take GCN as a backbone and performed grid search over the number of hidden layers, whether it has the structural embedding, the type of residual connection, and the type of normalization. For the number of hidden layers, we considered 2, 4, 8, 16, 32, and 64. For the types of residual connections, we include: (1) connection to the last layer~\citep{li2019deepgcns, li2018deeper}, (2) initial connection to the initial layer~\citep{chen2020simple, klicpera2018predict, zhang2020revisiting}, (3) dense connection to all preceding layers~\citep{li2019deepgcns, li2018deeper, li2020deepergcn, luan2019break}, and (4) jumping connection combining all the preceding layers only at the final graph convolutional layer~\citep{xu2018representation, liu2020towards}. For the types of normalizations, we grid search over batch normalization (BatchNorm)~\citep{ioffe2015batch}, pair normalization (PairNorm)~\citep{zhao2019pairnorm}, node normalization (NodeNorm)~\citep{zhou2020understanding}, mean normalization (MeanNorm)~\citep{yang2020revisiting}, and differentiable group normalization (GroupNorm)~\citep{zhou2020towards}.
For types of graph dropout methods, we include Dropout~\citep{srivastava2014dropout}, DropEdge~\citep{rong2020dropedge}, DropNode~\citep{dropedge2}, and LADIES~\citep{NEURIPS2019_91ba4a44}.

For the architecture design for Cold Brew's MLP, we conducted hyperparameter search over the number of hidden layers, the existence of residual connection, the hidden dimensions, and the optimizers. The number of hidden layers is searched over 2, 8, 16, and 32. The number of hidden dimensions is searched over 128 and 256. The optimizer is searched over (Adam(lr=0.001) Adam(lr=0.005), Adam(lr=0.02), SGD(lr=0.005))

% Our results show that the best architecture for these graph tasks is the two layer MLP without residual connection, with 128 hidden dimension in each hidden layers. We adopt this architecture to our student model.

For Label Propagation, we conducted hyperparameter search over the number of propagations, the propagation matrix type, and the mixing coefficient $\alpha$ \citep{CNS}. The number of propagations is searched over 10, 20, 50, 100, and 200. The propagation matrix type is searched over adjacency matrix and normalized Laplacian matrix. The mixing coefficient $\alpha$ is searched over 0.01, 0.1, 0.5, 0.9, and 0.99.

The best GCN, MLP, and LP configurations are reported in Tables~\ref{tab:bgcn}, \ref{tab:bmlp}, and \ref{tab:blp}, respectively.

\begin{table}[t]
\centering
\resizebox{0.7\textwidth}{!}{
\begin{tabular}{c|c | c  | c  | c c   }
\toprule[1.5pt]
\multirow{2}{*}{\textbf{\makecell[c]{Dataset}}} & \multicolumn{4}{c}{\textbf{\makecell{Best GCN}}} \\ %& \multicolumn{2}{c|}{\textbf{\makecell{Grid-Uni}}} & \multicolumn{2}{c}{\textbf{\makecell{NewYork}}}\\
\cmidrule(r){2-5}
 & num layers & whether has SE & residual type & normalization type  \\
% Dataset & GCN best configure & MLP Best configure & LP best configure  \\
   \hline
Cora  & 2 layer & has structural embedding & no residual & PairNorm \\
Citeseer  & 2 layer & has structural embedding & no residual & PairNorm \\
Pubmed  & 16 layer & has structural embedding & initial connection & GroupNorm \\
Arxiv  & 4 layer & has structural embedding & initial connection  & GroupNorm \\
Chameleon  & 2 layer & has structural embedding & initial connection  & BatchNorm \\
\bottomrule[1.5pt]
\end{tabular}}
\caption{Best GCN configurations.}
\label{tab:bgcn}
\end{table}

\begin{table}[t]
\centering
\resizebox{0.7\textwidth}{!}{
\begin{tabular}{c|c | c  | c  | c c   }
\toprule[1.5pt]
\multirow{2}{*}{\textbf{\makecell[c]{Dataset}}} & \multicolumn{4}{c}{\textbf{\makecell{Best MLP}}} \\ %& \multicolumn{2}{c|}{\textbf{\makecell{Grid-Uni}}} & \multicolumn{2}{c}{\textbf{\makecell{NewYork}}}\\
\cmidrule(r){2-5}
& hidden layers & residual connection & hidden dimensions & optimizer \\
 % & Num layers & whether has SE & residual type & normalization type  \\
% Dataset & GCN best configure & MLP Best configure & LP best configure  \\
   \hline
Cora  & 2 layer & no residual & 128 & Adam(lr=0.001) \\
Citeseer  & 4 layer & no residual & 128 & Adam(lr=0.001)\\
Pubmed  & 2 layer & no residual & 256 &  Adam(lr=0.02)\\
Arxiv  & 2 layer & no residual & 256 & Adam(lr=0.001)\\
Chameleon  & 2 layer & no residual & 256 & Adam(lr=0.001)\\
\bottomrule[1.5pt]
\end{tabular}
}
\caption{Best MLP configurations.}
\label{tab:bmlp}
\end{table}

\begin{table}[h]
\centering
\resizebox{0.7\textwidth}{!}{
\begin{tabular}{c|c | c  | c  c c   }
\toprule[1.5pt]
\multirow{2}{*}{\textbf{\makecell[c]{Dataset}}} & \multicolumn{4}{c}{\textbf{\makecell{Best LP}}} \\ %& \multicolumn{2}{c|}{\textbf{\makecell{Grid-Uni}}} & \multicolumn{2}{c}{\textbf{\makecell{NewYork}}}\\
\cmidrule(r){2-4}
& number of propagations & propagation matrix type & mixing coefficient & \\
% Dataset & GCN best configure & MLP Best configure & LP best configure  \\
   \hline
Cora  & 50 & Laplacian matrix & 0.1  \\
Citeseer  & 100 & Laplacian matrix & 0.01 \\
Pubmed  & 50 & Adjacency matrix & 0.5 \\
Arxiv  & 100 & Laplacian matrix & 0.5 \\
Chameleon  & 50 & Laplacian matrix & 0.1 \\
\bottomrule[1.5pt]
\end{tabular}}
\caption{Best Label Propagation configurations.}
\label{tab:blp}
\end{table}

\section{The performance on all splits of the data}
\label{appendix:all exps}
The performance evaluations on all splits are listed in Table~\ref{tab:fullexp}. The FCR evaluation on more datasets are presented in Figure~\ref{fig:hengfull}.
We hypothesize that a high FCR means that the graph does not add too much information for the task at hand. We indeed see evidence for this hypothesis in Table \ref{tab:fullexp}, where for the Pubmed dataset $(FCR \approx 77\% )$, the MLP-type models tend to outperform GNN-type models in all splits 
On the other hand, regardless of FCR, for almost all datasets, the MLP-type models outperform the GNN-type models on the isolation split, and a few on the tail split, while the GNN-type models are superior in other splits.

\begin{table}[h]
     \vspace{-0.5em}
    \centering
    \resizebox{0.9\textwidth}{!}{
    \begin{tabular}{l|lc|ccccc|cccccccccc}
    \toprule[1.5pt]
    \multirow{2}{*}{Splits} & \multicolumn{2}{c}{\multirow{2}{*}{Metrics/Models}} & \multicolumn{5}{|c|}{\bf Open-Source Datasets} & \multicolumn{4}{c}{\bf Proprietary Datasets} &  \\
    \cmidrule(r){4-8} \cmidrule(r){9-12}
    & & & Cora & Citeseer & Pubmed & Arxiv & Chameleon & E-comm1 & E-comm2  & E-comm3 & E-comm4 \\ 
    \midrule

    % __________________________________ exps here! __________________________________

      \multirow{6}{*}{Isolation}
     & \multirow{2}{*}{GNNs}  &  GCN 2 layers   & 58.02 & 47.09 & 71.50 & 44.51 & 57.28 & $-$ & $-$ & $-$ & $-$\\
     &   & GraphSAGE & 66.02 & 51.46 & 69.87 & 47.32 & 59.83 & +3.89 & +4.81 & +5.24 & +0.52\\
     \cmidrule(r){2-2}
     &  \multirow{2}{*}{MLPs}  &  Simple MLP  & 68.40 & \textbf{53.26} & 65.84 & 51.03 & 60.76 & +5.89 & +9.85 & +5.83 & +6.42\\
     &   &  GraphMLP  & 65.00 & 52.82 & 71.22 & 51.10 & \textbf{63.54 } & +6.27 & +9.46 & \textbf{+5.99 } & +7.37\\
     \cmidrule(r){2-2}
     &   \multirow{2}{*}{Cold Brew} &   GCN + SE 2 layers   & 58.37 & 47.78 & \textbf{73.85} & 45.20 & 60.13 & +0.27 & +0.76 & -0.50 & +1.22\\
     &   &  Student MLP  & \textbf{69.62} & 53.17 & 72.33 & \textbf{52.36} & 62.28 & \textbf{+7.56} & \textbf{+11.09} & +5.64 & \textbf{+9.05}\\

    \midrule  
     \multirow{6}{*}{Tail}

     & \multirow{2}{*}{GNNs}  &  GCN 2 layers   & 84.54 & \textbf{56.51} & 74.95 & 67.74 & 58.33 & $-$ & $-$ & $-$ & $-$\\
     &   & GraphSAGE & 82.82 & 52.77 & 73.07 & 63.23 & 61.26 & -3.82 & -3.07 & -2.87 & -6.42\\
     \cmidrule(r){2-2}
     &  \multirow{2}{*}{MLPs} &  Simple MLP  & 70.76 & 54.85 & 67.21 & 52.14 & 50.12 & -0.37 & +1.74 & -0.13 & -0.45\\
     &   &  GraphMLP  & 70.09 & 55.56 & 71.45 & 52.40 & 52.84 & -0.33 & +1.64 & \textbf{+1.27 }& +0.80\\
     \cmidrule(r){2-2}
     &  \multirow{2}{*}{Cold Brew}   &   GCN + SE 2 layers   &\textbf{ 84.66 }& 56.32 & \textbf{75.33} & \textbf{68.11} & \textbf{60.80} & \textbf{+0.85} & +0.44 & -0.60 & +1.10\\
     &   &  Student MLP  & 71.80 & 54.88 & 72.54 & 53.24 & 51.36 & +0.32 & \textbf{+3.09 }& -0.18 & \textbf{+2.09}\\

      \midrule 
     \multirow{6}{*}{Head}  
     & \multirow{2}{*}{GNNs}   &  GCN 2 layers   & 88.68 & 80.37 & 85.79 & 73.35 & 67.49 & $-$ & $-$ & \textbf{$-$ }& $-$\\
     &   & GraphSAGE & 87.75 & 74.81 & 86.94 & 70.85 & 62.08 & -4.26 & -4.17 & -3.50 & -7.46\\
     \cmidrule(r){2-2}
     &  \multirow{2}{*}{MLPs} &  Simple MLP  & 74.33 & 72.00 & 89.00 & 56.34 & 60.82 & -16.74 & -18.10 & -16.73 & -16.51\\
     &   &  GraphMLP  & 72.45 & 69.83 & 89.00 & 56.65 & 62.44 & -15.96 & -18.08 & -15.33 & -15.41\\
     \cmidrule(r){2-2}
     &  \multirow{2}{*}{Cold Brew}   &   GCN + SE 2 layers   & \textbf{89.39 } & \textbf{80.76} & 87.83 &\textbf{74.01} & 70.56 & \textbf{+1.11} & \textbf{+0.47 }& -0.39 & \textbf{+1.28}\\
     &   &  Student MLP  & 74.53 & 72.33 & \textbf{90.33} & 57.41 & 61.28 & -15.28 & -17.42 & -17.02 & -15.41\\

     \midrule 
     \multirow{6}{*}{Overall} 
     
     &  \multirow{2}{*}{GNNs}  &  GCN 2 layers   & 84.89 & 70.38 & 78.18 & 71.50 & 59.30 & $-$ & \textbf{$-$} & \textbf{$-$} & $-$\\
     &   & GraphSAGE & 80.90 & 66.21 & 76.73 & 68.33 & 70.02 & -3.09 & -3.86 & -2.58 & -5.48\\
     \cmidrule(r){2-2}
     & \multirow{2}{*}{MLPs}  &  Simple MLP  & 69.02& 56.59& 73.51& 54.89& 58.65 & -12.69 & -12.86 & -12.68 & -13.16\\
     &   &  GraphMLP  & 71.87 & 68.22 & 82.03 & 53.81 & 57.67 & -12.26 & -12.01 & -10.80 & -11.41\\
     \cmidrule(r){2-2}
     &   \multirow{2}{*}{Cold Brew}  &   GCN + SE 2 layers   & \textbf{86.96} & \textbf{72.44} & 79.03 & \textbf{71.92} & \textbf{68.51} & \textbf{+0.65} & -0.24 & -0.77 & \textbf{+1.43}\\
     &   &  Student MLP  & 72.36 & 67.54 &\textbf{ 82.00} & 54.94 & 59.07 & -11.25 & -11.51 & -11.55 & -11.21\\

    % __________________________________ exps here! __________________________________

    \bottomrule[1.5pt]
    \end{tabular}}
    \caption{The performance comparisons on all splits of different datasets.}
    \label{tab:fullexp}
    \vspace{1em}
    \end{table}

\begin{table}
     \vspace{-0.5em}
     \centering
    \resizebox{0.9\columnwidth}{!}{
    \begin{tabular}{c|c  c   c c c cccccc  }
    % \centering
    \toprule[1.5pt]
    % \multirow{2}{*}{\textbf{\makecell[c]{Dataset}}} & \multicolumn{4}{c}{\textbf{\makecell{d}}} \\ %& \multicolumn{2}{c|}{\textbf{\makecell{Grid-Uni}}} & \multicolumn{2}{c}{\textbf{\makecell{NewYork}}}\\
    % \cmidrule(r){2-5}
     & Cora & Citeseer & Pubmed & Arxiv & Cham. & Squ. & Actor & Cornell & Texas & Wisconsin   \\
    % Dataset & GCN best configure & MLP Best configure & LP best configure  \\
       \hline
    GNN  & 86.96& 72.44& 75.96& 71.54& 68.51& 31.95& 59.79& 65.1 & 61.08& 81.62 \\

    MLP  & 69.02& 56.59& 73.51& 54.89& 58.65& 38.51& 37.93& 86.26& 83.33& 85.42 \\
    Label Propagation  &  78.18& 45.00  & 67.8 & 68.26& 41.01& 22.85& 29.69& 32.06& 52.08&  40.62 \\

    FCR \% & 32.86 \% &  63.39 \% &  76.91\% &  16.45\% &  73.61\% & 141.91\% &  57.93\% & 139.04\% & 171.2 \% & 108.48 \% \\
    $\beta(\hG)$ \% & 83\% & 71\% & 79\% & 68\% & 25\% & 22\% & 24\% & 11\% & 6\% & 16\% \\
       \hline
    $head-tail (GNN)$ &  4.44&  23.98&  11.71&   5.9 &   0.24&  -6.51&   2.22&  -4.37& -11.26& -33.92 \\
    $head-isolation (GNN)$ & 31.01&  33.09&  15.21&  28.81&   1.55&  -4.85&  22.61& -18.68& -24.62& -29.23  \\
    \bottomrule[1.5pt]
    \end{tabular}
    }
    \caption{Top part: FCR and its components. The $\beta$ metric is added as a reference. Bottom part: the performance difference of GNN on the head/tail and head/isolation splits.}
     \vspace{-0.5em}
    \label{fig:hengfull}
    \end{table}

\section{Visualizing the learned embeddings}

Figure~\ref{fig:viz} visualizes the last-layer embeddings of different models after t-SNE dimensionality reduction. In the figure, colors denotes node labels and all nodes are marked as dots, with isolation subset nodes additionally marked with \textit{x}'s and the tail subset additionally marked with triangles. Although the GCN model did a decent job in separating different classes, a significant portion of the tail and isolation nodes fall into wrong class clusters. Cold Brew's MLP is more discriminative in the tail and isolation splits.

\def\subw{2.4in}
    % need to : \usepackage{subfigure}
    \begin{figure}[b]
     \vspace{-1em}
        \centering
        \subfigure{
            \includegraphics[width=\subw]{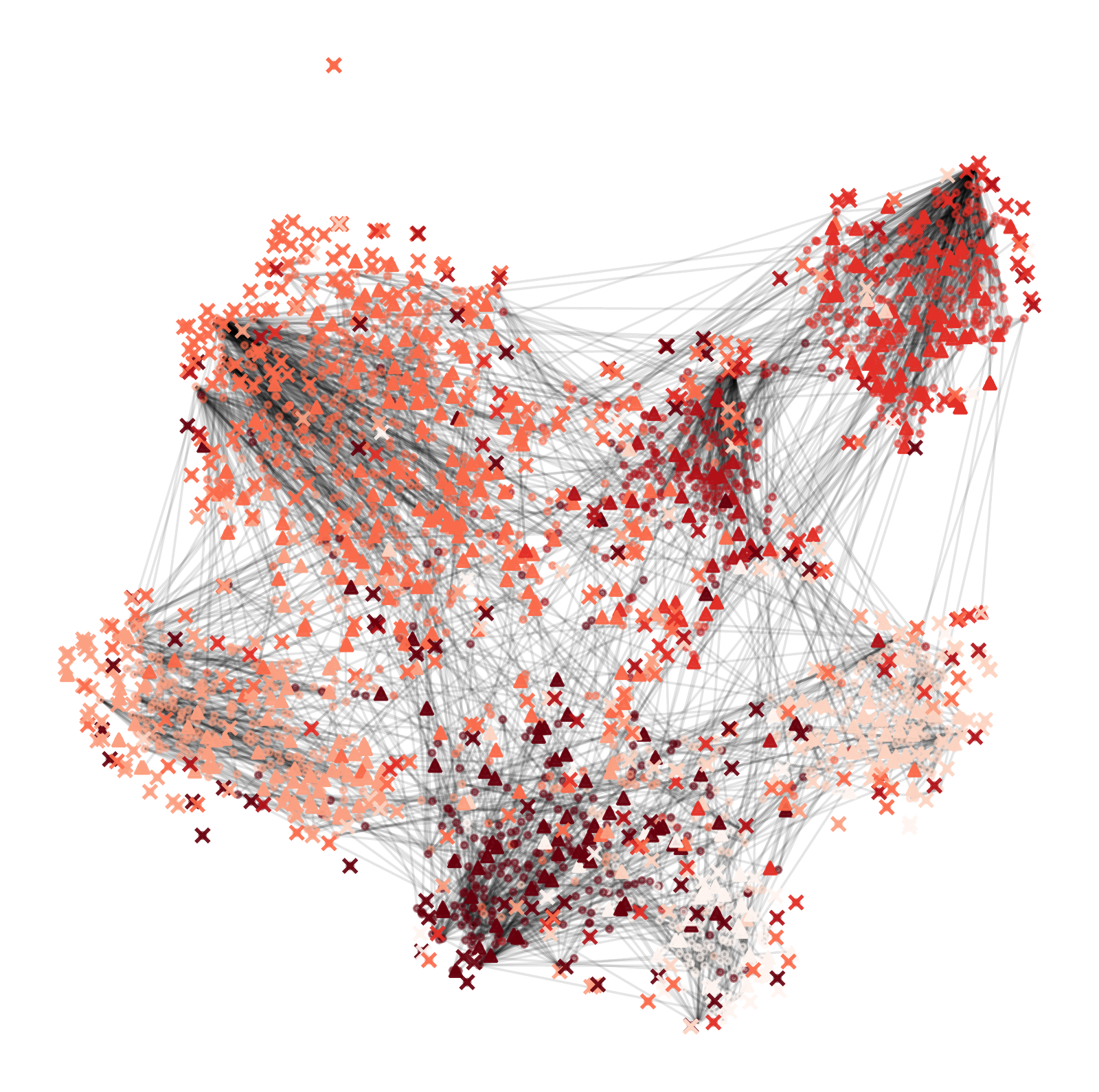}
        }
        \subfigure{
            \includegraphics[width=\subw]{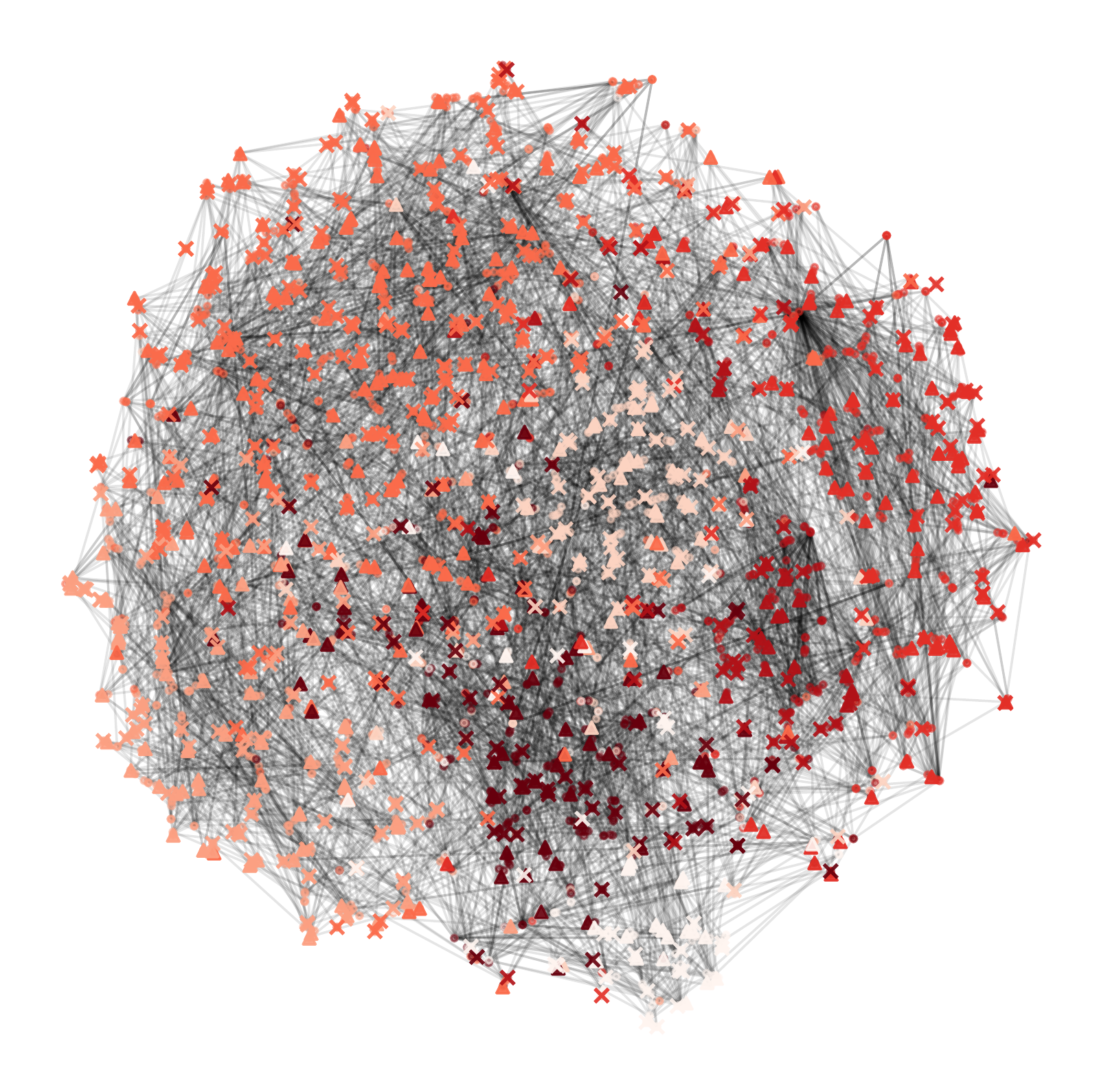}
        }
        \subfigure{
            \includegraphics[width=\subw]{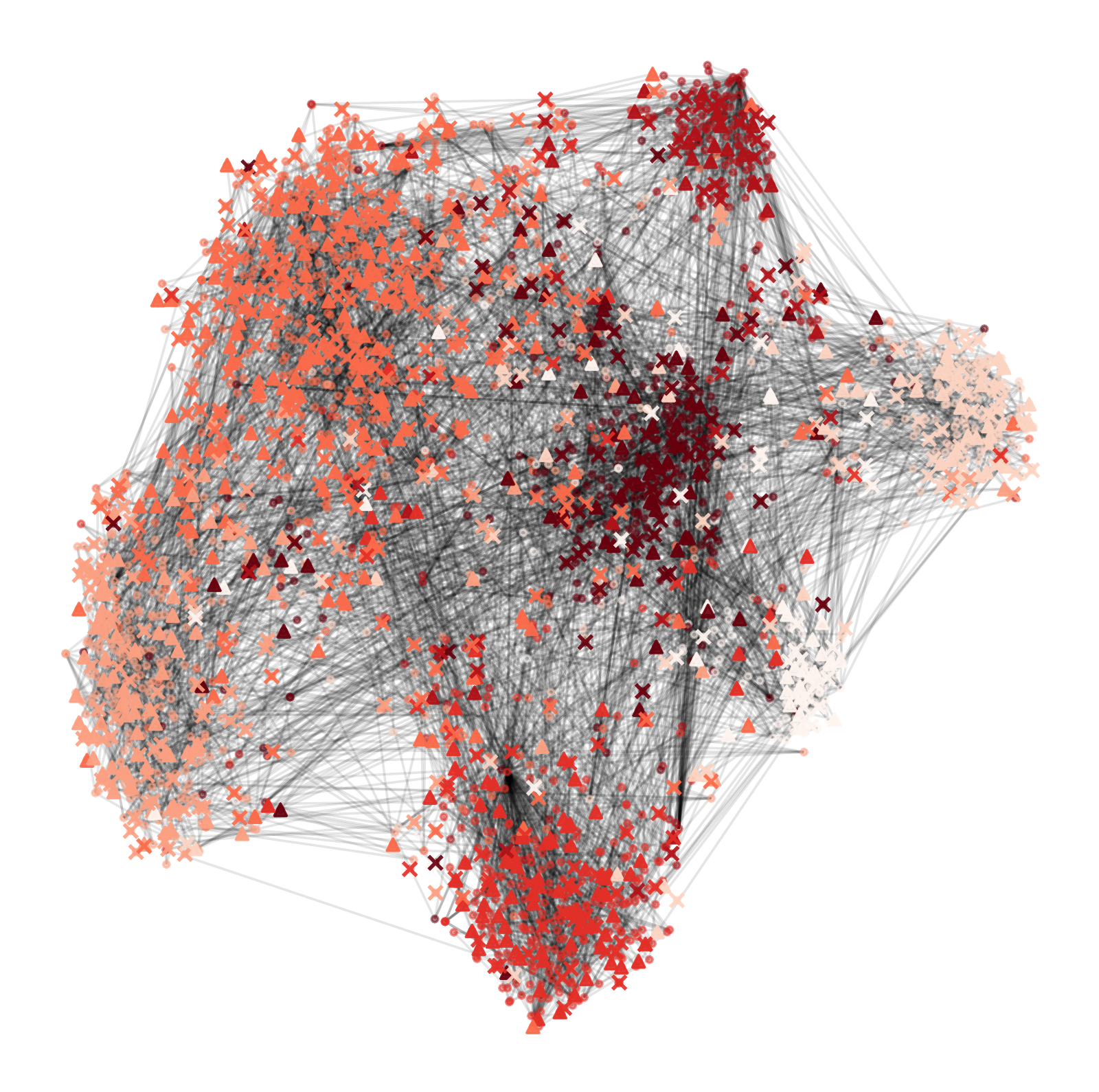}
        }
        \subfigure{
            \includegraphics[width=\subw]{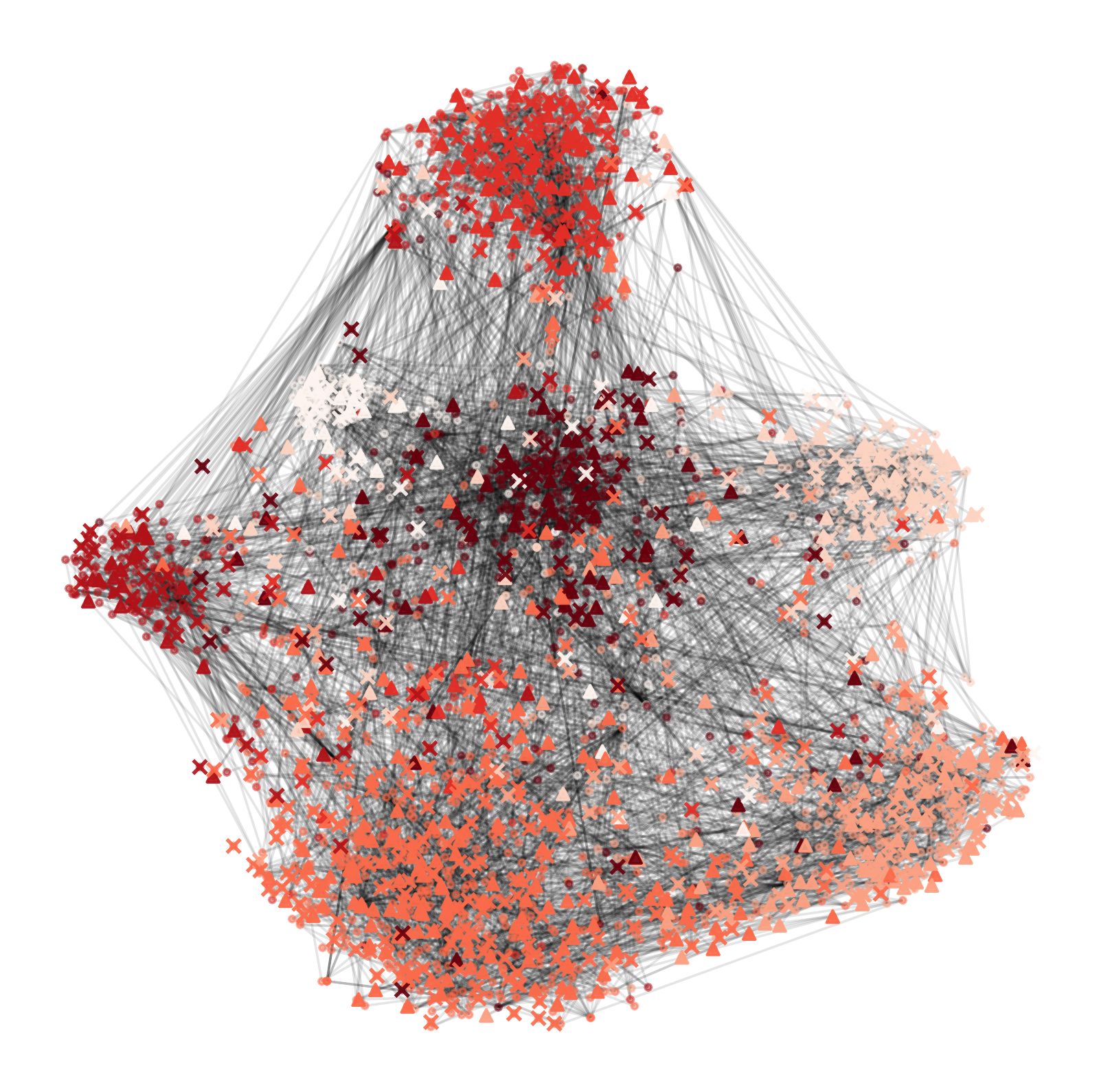}
        }
        \vspace{-1em}
        \caption{Top two subfigures: the last-layer embeddings of GCN and Simple MLP. Bottom two subfigures: the last-layer embeddings of GraphMLP and Cold Brew's student MLP. All embeddings are projected to 2D with t-SNE. Cold Brew's MLP has the fewest isolated nodes that are misplaced into wrong clusters.}
         \vspace{-1em}
        \label{fig:viz}
    \end{figure}

\end{document}